%% file: main.tex
\documentclass[10pt,twocolumn,letterpaper]{article}
\usepackage[accsupp]{axessibility}
\usepackage{cvpr}              

\usepackage{booktabs}
\usepackage{makecell}
\usepackage[table]{xcolor}
\usepackage{tcolorbox}
\definecolor{datacellgreen}{HTML}{D9EAD3}
\definecolor{datacellpink}{HTML}{FFD1DC}
\input{preamble}
\definecolor{cvprblue}{rgb}{0.21,0.49,0.74}
\usepackage[pagebackref,breaklinks,colorlinks,allcolors=cvprblue]{hyperref}

\def\paperID{CV4Clinic 2026-66} 
\def\confName{CVPR}
\def\confYear{2026}

\title{RobustMedSAM: Degradation-Resilient Medical Image Segmentation via Robust Foundation Model Adaptation}

\author{
Jieru Li, Matthew Chen, Micky C. Nnamdi, J. Ben Tamo, Benoit L. Marteau, May D. Wang\thanks{Corresponding author.} \\
Georgia Institute of Technology \\
\tt\small{\{jli3545, mchen439, mnnamdi3, jtamo3, bmarteau3, maywang\}@gatech.edu}
}

\begin{document}
\maketitle
\input{sec/0_abstract}    
\input{subsection/1-intro}
\input{subsection/2-related}
\input{subsection/3-method}
\input{subsection/4-results}
\input{subsection/5-conclusion}
{
    \small
    \bibliographystyle{ieeenat_fullname}
    \bibliography{main}
}
\input{subsection/6-supplementary}

\end{document}

%% file: sec/0_abstract.tex
\begin{abstract}
Medical image segmentation models built on Segment Anything Model (SAM) achieve strong performance on clean benchmarks, yet their reliability often degrades under realistic image corruptions such as noise, blur, motion artifacts, and modality-specific distortions. Existing approaches address either medical-domain adaptation or corruption robustness, but not both jointly. In SAM, we find that these capabilities are concentrated in complementary modules: the image encoder preserves medical priors, while the mask decoder governs corruption robustness. Motivated by this observation, we propose \textbf{RobustMedSAM}, which adopts module-wise checkpoint fusion by initializing the image encoder from MedSAM and the mask decoder from RobustSAM under a shared ViT-B architecture. We then fine-tune only the mask decoder on 35 medical datasets from MedSegBench, spanning six imaging modalities and 12 corruption types, while freezing the remaining components to preserve pretrained medical representations. We additionally investigate an SVD-based parameter-efficient variant for limited encoder adaptation. Experiments on both in-distribution and out-of-distribution benchmarks show that RobustMedSAM improves degraded-image Dice from 0.613 to 0.719 (+0.106) over SAM, demonstrating that structured fusion of complementary pretrained models is an effective and practical approach for robust medical image segmentation.
\end{abstract}

%% file: subsection/1-intro.tex
\section{Introduction}
\label{sec:intro}

Medical image segmentation is critical to many clinical workflows, yet its reliability often degrades in practice due to acquisition noise, motion artifacts, blur, and modality-specific distortions~\cite{zhang2023enhancing, noise, chen2024robustsam}. These degradations are intrinsic to routine medical imaging and can substantially reduce segmentation quality, but most existing methods are developed and evaluated primarily on clean data.

The Segment Anything Model (SAM)~\cite{kirillov2023segment} introduced a general promptable segmentation framework, and MedSAM~\cite{ma2024segment} adapted it to the medical domain through large-scale supervised fine-tuning. While MedSAM performs well on clean in-domain data, it does not explicitly address image degradation. In parallel, RobustSAM~\cite{chen2024robustsam} improves SAM robustness to corruption by enhancing the mask decoder, but it is designed for natural images and lacks medical-domain priors. Recent work has also explored universal medical segmentation via one-shot prompting~\cite{wu2024one} and parameter-efficient adaptation of SAM's encoder through cross-block orchestration~\cite{peng2024parameter}, yet none of these methods jointly target degradation robustness and medical-domain adaptation. Zhang et al.~\cite{zhang2024improving} further showed that SAM's generalization degrades significantly under distribution shift, including on corrupted and medical images, reinforcing the need for explicit robustness mechanisms.
\begin{figure*}
\centering
\includegraphics[width=\textwidth]{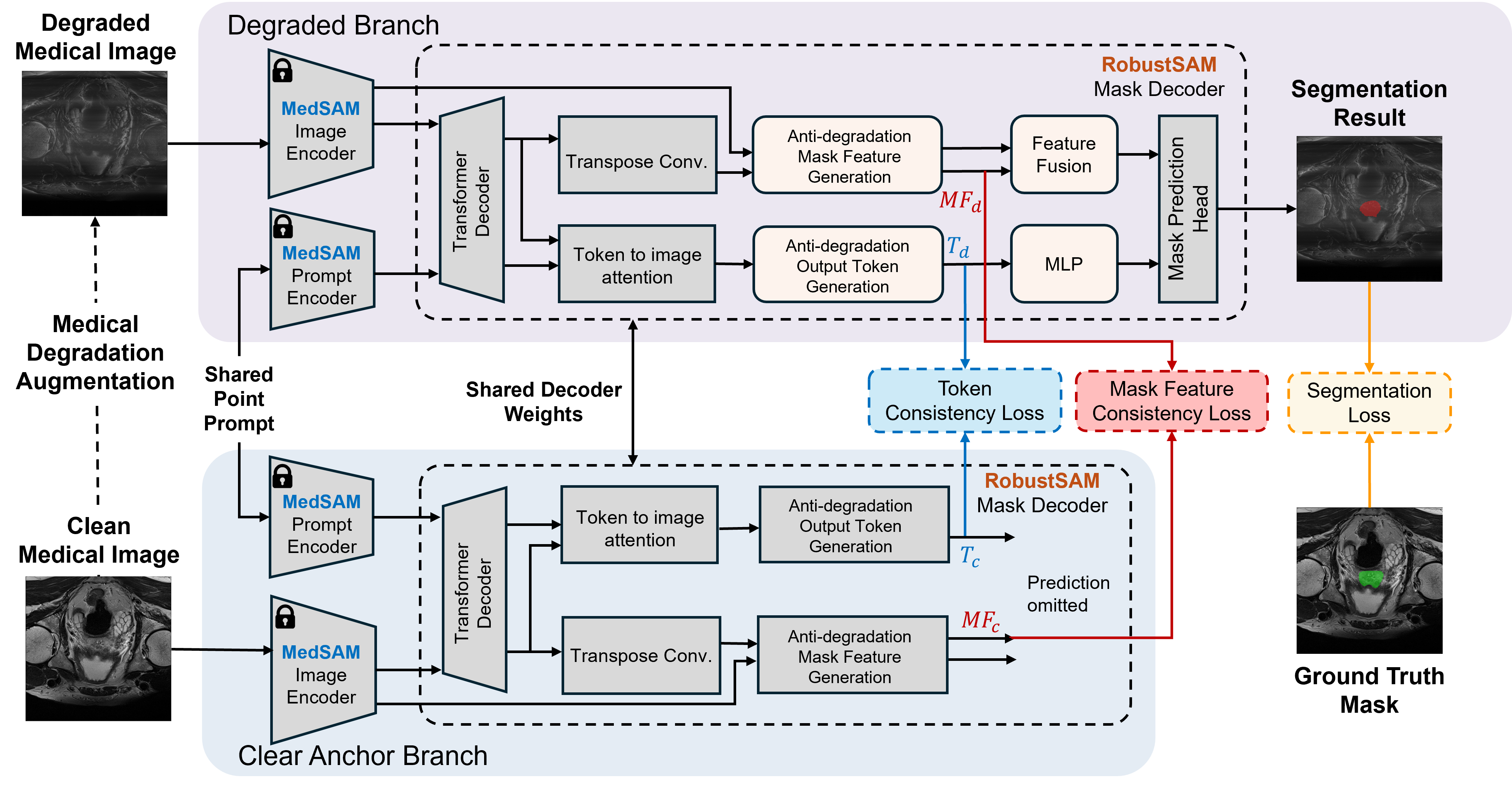}
\caption{
\textbf{Overview of RobustMedSAM.}
During training, each medical image is paired with a degraded counterpart generated by medical degradation augmentation. Both the clean and degraded images are processed using the same frozen MedSAM image encoder and prompt encoder. The resulting features are decoded by a shared robust decoder initialized from RobustSAM. Finetuned modules are highlighted in color, while frozen modules are shown in gray. At inference time, RobustMedSAM requires only a single input image without any clean reference.
}
\label{fig:robustmedsam_pipeline}
\end{figure*}
Our key observation is that these two capabilities are complementary and largely concentrated in different modules: MedSAM provides a medically adapted image encoder, while RobustSAM provides a corruption-resilient mask decoder. This modularity motivates a direct fusion strategy rather than training a new model from scratch. A natural question arises: can module-wise checkpoint fusion, initializing different components from models trained for complementary tasks, effectively compose these capabilities?

To investigate this, we propose \textbf{RobustMedSAM}, a degradation-resilient medical segmentation framework that combines MedSAM and RobustSAM through mixed checkpoint initialization. We initialize the image encoder from MedSAM and the mask decoder from RobustSAM, then fine-tune only the mask decoder on 35 medical segmentation datasets spanning six imaging modalities and 12 corruption types from MedSegBench~\cite{kucs2024medsegbench}, while keeping the remaining components frozen. We also investigate an SVD-based parameter-efficient variant for limited encoder adaptation.

Extensive experiments show that RobustMedSAM consistently improves robustness under image degradation. On MedSegBench, it improves degraded-image Dice from 0.613 to 0.719 over SAM, and shows improvements on most out-of-distribution datasets, with limited gains on topology-sensitive structures. These results suggest that structured fusion of complementary pretrained models is an effective and practical solution for robust medical image segmentation. Our contributions are summarized as follows:
\begin{itemize}
    \item We demonstrate that module-wise checkpoint fusion, initializing the encoder from a medically adapted model (MedSAM) and the decoder from a corruption-robust model (RobustSAM), effectively composes medical-domain adaptation with corruption robustness, outperforming each parent model on average across 35 datasets spanning six modalities.

    \item We show that decoder-only fine-tuning with clean--degraded consistency training is sufficient for robust adaptation, preserving clean-image accuracy while improving degraded-image Dice by $+0.106$ over SAM, and that SVD-based encoder adaptation introduces an unfavorable robustness--clean trade-off.


    \item Through evaluation across six modalities, 12 degradation types, and multiple prompt settings, we characterize when module-wise fusion succeeds, most clearly under medical-specific corruptions and point-based prompting, and where it remains limited, including topology-sensitive structures and box-prompt settings.
    
\end{itemize}

%% file: subsection/2-related.tex
\section{Related Work}
\label{sec:related}

\noindent\textbf{Foundation Models for Segmentation.}
SAM~\cite{kirillov2023segment} established the prompt-based segmentation paradigm, demonstrating strong zero-shot generalization on natural images using a ViT encoder, a prompt encoder, and a mask decoder trained on over one billion masks. However, SAM's performance degrades on out-of-distribution domains such as medical imaging, where texture, contrast, and noise characteristics differ substantially from those in natural photographs. Beyond the original SAM, recent studies have explored improving the efficiency and deployability of promptable segmentation foundation models (e.g., EfficientSAM~\cite{xiong2024efficientsam}), highlighting the rapid evolution of the SAM ecosystem. MedSAM~\cite{ma2024segment} addressed this by fine-tuning SAM on 1.57 million medical image--mask pairs across 10+ modalities, achieving strong benchmark performance. Nevertheless, MedSAM was trained exclusively with bounding-box prompts on clean, tightly cropped inputs; conditions that are difficult to guarantee in clinical practice.

\noindent\textbf{Medical Adaptations of SAM.}
Several works have sought to improve SAM's applicability to medical imaging along different axes. AutoSAM~\cite{shaharabany2023autosam} replaces the manual prompt encoder with a learned module that generates prompts directly from the input image, enabling automatic segmentation while keeping SAM's weights frozen. Object-detection-based pipelines such as YOLOv8+SAM~\cite{Pandey_2023_ICCV} and MedLSAM~\cite{medlsam} automate prompt generation by first localizing target structures via detection networks, then passing the predicted bounding boxes to SAM. For domain-specific adaptation, SurgicalSAM~\cite{yue2024surgicalsam} introduces efficient class-promptable tuning for surgical instrument segmentation, whereas SAM-DA~\cite{tejero2025sam} inserts lightweight adapter modules into SAM's decoder for parameter-efficient medical domain alignment. In parallel, alternative medical adaptations have also been proposed, including hierarchical decoding and prompt-reduced variants (e.g., H-SAM~\cite{cheng2024unleashing}),
as well as multi-source training strategies that handle partially labeled datasets across modalities~\cite{chen2024versatile}. UR-SAM~\cite{zhang2023enhancing} takes an orthogonal approach, improving reliability by estimating segmentation uncertainty from perturbed prompts and rectifying the output mask in high-uncertainty regions. While these methods address prompt automation, domain shift, or prediction uncertainty, none explicitly target robustness to image degradation.

\noindent\textbf{Robustness to Image Degradation.}
Image corruptions, including noise, blur, compression artifacts, and modality-specific distortions, are pervasive in clinical imaging yet often overlooked by medical segmentation models. Robustness to corruptions has been systematically studied in semantic segmentation, including large-scale benchmarking efforts revealing substantial performance degradation~\cite{kamann2020benchmarking}. More targeted efforts show that even a single corruption family, such as motion blur, can destabilize segmentation, motivating corruption-aware augmentation and training strategies~\cite{rajagopalan2023improving}. In the natural image domain, RobustSAM~\cite{chen2024robustsam} demonstrated that SAM can be made resilient to diverse degradations by augmenting its mask decoder with four lightweight modules trained on 688K synthetically corrupted images, without sacrificing promptability or zero-shot capability. Frequency-domain perspectives separating semantic content from nuisance factors have also been explored for domain shift and robustness (e.g., FDA~\cite{yang2020fda}); 
Nam et al.~\cite{nam2024modality} further showed that multi-frequency attention integrated with multi-scale features achieves modality-agnostic domain generalization across six medical imaging modalities and fifteen datasets, underscoring the importance of frequency-aware representations for robust medical segmentation.

Despite these advances, no prior work jointly addresses medical domain adaptation with degradation robustness in a unified framework. RobustSAM handles corruptions but lacks medical domain knowledge; MedSAM encodes medical features but is fragile to variations in image quality. Our work bridges this gap by fusing the complementary strengths of both models.

%% file: subsection/3-method.tex
\section{Method}
\label{sec:method}
Our central hypothesis is that medical priors and degradation robustness are largely encoded in distinct functional components of SAM-family models, the encoder and decoder, respectively, and can therefore be composed by allocating pretrained parameters module-wise rather than uniformly fine-tuning the full network. We test this through RobustMedSAM, a medical-domain-preserving adaptation framework for segmentation under realistic image corruptions.

Concretely, we initialize the image encoder from MedSAM to retain modality-specific feature statistics and anatomical representations learned from large-scale medical supervision. We initialize the mask decoder from RobustSAM, as corruption robustness is primarily driven by decoder-side calibration and token refinement. This separation preserves medical priors while providing a robust decoding head from the start, reducing interference between domain adaptation and robustness learning.

RobustMedSAM comprises three components: (1) module-wise mixed initialization, (2) robustness-aware decoding trained with clean--degraded cross-branch alignment, and (3) an optional SVD-based, parameter-efficient encoder adaptation.

\subsection{Preliminaries}
\label{sec:prelim}

\noindent\textbf{SAM}~\cite{kirillov2023segment} consists of an image encoder $\mathcal{E}$, a prompt encoder $\mathcal{P}$, and a mask decoder $\mathcal{D}$.
Given an image $\mathbf{x}\in\mathbb{R}^{H\times W\times 3}$ and point prompts $\{(p_i,l_i)\}_{i=1}^{K}$ where $p_i\in\mathbb{R}^2$ and $l_i\in\{0,1\}$, SAM predicts
\begin{equation}
\hat{\mathbf{m}}=\mathcal{D}\!\left(\mathcal{E}(\mathbf{x}),\,\mathcal{P}(\{(p_i,l_i)\}_{i=1}^{K})\right),\quad
\hat{\mathbf{m}}\in[0,1]^{H\times W}.
\end{equation}

\noindent\textbf{MedSAM}~\cite{ma2024segment} adapts SAM to medical domains by fine-tuning on large-scale medical image--mask pairs. It primarily assumes clean inputs and bounding-box prompts, and does not explicitly target degradation robustness.

\noindent\textbf{RobustSAM}~\cite{chen2024robustsam} strengthens SAM decoding under corruptions by augmenting the decoder with anti-degradation modules and enforcing degradation-invariant decoding via feature/token consistency regularization.

\subsection{Model Architecture}
\label{sec:arch}

An overview of the proposed RobustMedSAM framework is illustrated in Fig.~\ref{fig:robustmedsam_pipeline}.

\noindent\textbf{Mixed Initialization.}
RobustMedSAM initializes the encoder and prompt encoder from MedSAM and the robust decoder from RobustSAM:
\begin{equation}
\theta_{\mathcal{E}}\leftarrow\theta_{\mathcal{E}}^{\text{MedSAM}},\quad
\theta_{\mathcal{P}}\leftarrow\theta_{\mathcal{P}}^{\text{MedSAM}},\quad
\theta_{\mathcal{D}}\leftarrow\theta_{\mathcal{D}}^{\text{RobustSAM}}.
\end{equation}
Although the backbones are dimensionally compatible, mixed initialization may introduce an encoder--decoder feature mismatch, which we mitigate through decoder-side alignment regularizers during training (Sec.~\ref{sec:training}).

\noindent\textbf{Clean--Degraded Pair Training.}
For each sample, we construct a clean--degraded pair $(\mathbf{x}_c,\mathbf{x}_d)$ from the same underlying image, with a shared ground-truth mask $\mathbf{m}$:
\begin{equation}
\mathbf{x}_d=\mathcal{T}(\mathbf{x}_c),\quad \mathcal{T}\sim\mathcal{S}.
\end{equation}
where $\mathcal{S}$ denotes a predefined degradation distribution over corruption types and severities, and $\mathcal{T}$ is a sampled degradation operator applied to $\mathbf{x}_c$.

Both views share the same encoder and the same prompt tokens $\mathbf{p}$:
\begin{equation}
\mathbf{f}_c=\mathcal{E}(\mathbf{x}_c),\quad
\mathbf{f}_d=\mathcal{E}(\mathbf{x}_d),\quad
\mathbf{p}=\mathcal{P}(\{(p_i,l_i)\}_{i=1}^{K}).
\end{equation}
The robust decoder predicts masks for both branches:
\begin{equation}
\hat{\mathbf{m}}_c=\mathcal{D}(\mathbf{f}_c,\mathbf{p}),\quad
\hat{\mathbf{m}}_d=\mathcal{D}(\mathbf{f}_d,\mathbf{p}).
\end{equation}
During training, the clean branch serves as a semantic anchor for decoder alignment, while segmentation supervision is applied only to the degraded branch to directly optimize robustness under corruption. We choose to train the Anti-degradation Mask Feature Generation and Anti-degradation Output Token Generation modules adapted from RobustSAM. Detailed architectures of these modules are provided in the supplementary material~\ref{sec:arch_detail}.

\subsection{Training Objective}
\label{sec:training}

We freeze $\mathcal{E}$ and $\mathcal{P}$ and optimize only the robust decoder $\mathcal{D}$.
We treat each clean--degraded pair as a positive pair that shares anatomy but differs in corruption. This paired formulation is used only during training; during inference, RobustMedSAM operates on a single image without requiring any clean reference.

\noindent\textbf{Segmentation Loss.}
We supervise only $\hat{\mathbf{m}}_d$ using Dice and Focal losses:
\begin{equation}
\mathcal{L}_{\text{seg}}
=
\alpha\,\mathcal{L}_{\text{dice}}(\hat{\mathbf{m}}_d,\mathbf{m})
+
\beta\,\mathcal{L}_{\text{focal}}(\hat{\mathbf{m}}_d,\mathbf{m}).
\end{equation}

\noindent\textbf{Mask Feature Consistency.}
Let $\mathbf{MF}_c$ and $\mathbf{MF}_d$ denote intermediate mask features for the clean and degraded branches, respectively. We enforce
\begin{equation}
\mathcal{L}_{\text{mfc}}
=
\mathrm{MSE}(\mathbf{MF}_d,\mathbf{MF}_c).
\end{equation}
This loss encourages degradation-invariant features and improves segmentation robustness.

\noindent\textbf{Token Consistency.}
Let $\mathbf{t}_c,\mathbf{t}_d\in\mathbb{R}^{C}$ denote the refined output tokens for clean and degraded branches. We impose
\begin{equation}
\mathcal{L}_{\text{tc}}
=
\mathrm{MSE}(\mathbf{t}_d,\mathbf{t}_c).
\end{equation}
This constraint encourages the refined token to remain consistent with clear-image representations.

\noindent\textbf{Total Loss.}
\begin{equation}
\mathcal{L}
=
\mathcal{L}_{\text{seg}}
+
\lambda_1 \mathcal{L}_{\text{mfc}}
+
\lambda_2 \mathcal{L}_{\text{tc}}.
\end{equation}
We set $\alpha{=}20$, $\beta{=}1$, $\lambda_1{=}100$, and $\lambda_2{=}2$ to balance term magnitudes; a detailed justification is provided in the \textbf{Supplementary Material}.


\subsection{SVD-Based Encoder Adaptation}
\label{sec:svd}

Freezing the encoder preserves medical-domain priors but may limit adaptation under severe distribution shift. As a lightweight alternative to full fine-tuning, we explore a structured parameter-efficient adaptation based on singular value decomposition (SVD).

Given a convolutional weight matrix $\mathbf{W} \in \mathbb{R}^{C_{\text{out}} \times C_{\text{in}}k^2}$, we decompose it as:
\begin{equation}
\mathbf{W} = \mathbf{U} \boldsymbol{\Sigma} \mathbf{V}^\top.
\end{equation}
We freeze the orthogonal bases $\mathbf{U}$ and $\mathbf{V}$ inherited from MedSAM and update only the singular values $\boldsymbol{\Sigma}$. During training, the weight is reconstructed as:
\begin{equation}
\mathbf{W}' = \mathbf{U} \boldsymbol{\Sigma}' \mathbf{V}^\top.
\end{equation}

This strategy adjusts the relative scaling of existing feature directions without modifying the learned feature subspace. Compared with full encoder fine-tuning, it introduces only
\begin{equation}
r = \min(C_{\text{out}}, C_{\text{in}}k^2)
\end{equation}
additional trainable parameters (less than 0.1\% of encoder parameters in our ViT-B configuration), and serves as an optional robustness-oriented refinement. We evaluate this variant in Sec.~\ref{sec:ablation}.

\subsection{MedSegBench Degradation Protocol}
\label{sec:benchmark}
We evaluate on MedSegBench~\cite{kucs2024medsegbench}, which contains 35 datasets spanning six imaging modalities under 12 corruption types. Corruptions reflect modality-specific acquisition artifacts (e.g., Rician noise in MRI, speckle in ultrasound) alongside modality-agnostic perturbations (Gaussian noise, blur, brightness, contrast) applied uniformly across all datasets. Detailed mappings are provided in Table~\ref{tab:modality_degradations}.




\begin{table}
\centering
\caption{Degradation types applied to each imaging modality}
\label{tab:modality_degradations}
\resizebox{\linewidth}{!}{
\begin{tabular}{lcc}
\hline
\textbf{Modality} & \textbf{Applied Degradations} \\
\hline
Ultrasound & Speckle noise, Salt-and-pepper noise \\
MRI & Rician noise, Rayleigh noise, Step-wise motion artifact \\
CT & Poisson noise, Step-wise motion artifact \\
X-Ray, Microscopy & Poisson noise \\
Endoscopy & Motion blur, Zoom blur, Compression artifacts \\
Fundus & ISO noise, Zoom blur, Compression artifacts \\
OCT & Speckle noise \\
Pathology & ISO noise, Color jitter \\
Nuclei / NuclearCell & Salt-and-pepper noise, Color jitter \\
Dermoscopy & Color jitter, Compression artifacts \\
\hline
All modalities & Gaussian noise, Gaussian blur, Contrast variation, Brightness shift \\
\hline
\end{tabular}
}
\end{table}





%% file: subsection/4-results.tex
\section{Experiments}
\label{sec:experiments}

\subsection{Experimental Setup}
\label{sec:setup}

\noindent\textbf{Datasets.}
We use MedSegBench~\cite{kucs2024medsegbench}, which covers 35 medical image segmentation datasets across six modalities and includes 12 synthetic corruption types (Sec.~\ref{sec:benchmark}). We report in-distribution results on six representative datasets (ISIC~2016, BUSI, PROMISE12, UWaterloo Skin Cancer, BriFiSeg, and AbdomenUS), and evaluate out-of-distribution generalization on six held-out datasets (CHASE\_DB1, DRIVE, ISIC~2018, Kvasir, PanDental, and DCA1).

\noindent\textbf{Baselines.}
We compare against \textbf{SAM}~\cite{kirillov2023segment}, \textbf{MedSAM}~\cite{ma2024segment}, and \textbf{RobustSAM}~\cite{chen2024robustsam}. Unless otherwise stated, all methods are evaluated with $K{=}3$ foreground point prompts. The foreground points are randomly sampled from the ground-truth mask for each image. Although MedSAM was originally trained with box prompts, we report its point-prompt performance for a unified evaluation; point prompts represent a more demanding setting that avoids reliance on tight bounding boxes. Box-prompt results, where MedSAM performs substantially better, are provided in the Supplementary (Table~\ref{tab:bbox_main}) for completeness. We further analyze sensitivity to prompt type and the number of point prompts in Sec.~\ref{sec:prompt_sensitivity}.

\noindent\textbf{Implementation Details.}
All models use a ViT-B backbone. We freeze the MedSAM image encoder and prompt encoder and train the mask decoder initialized from RobustSAM for 10 epochs using Adam (learning rate: $5\times10^{-4}$), batch size 4. For the SVD variant, we additionally optimize the encoder neck convolution's singular values (Sec.~\ref{sec:svd}). Training is performed on a single NVIDIA H200 GPU, with all images resized to $512\times512$.

\noindent\textbf{Metrics.} We report Dice coefficient, intersection over union (IoU), and normalized surface distance (NSD, $\tau{=}2$\,px). Formal definitions are provided in the Supplementary Material.

\subsection{Main Results}
\label{sec:main_results}

Table~\ref{tab:main} summarizes segmentation performance on degraded test images. For each dataset, we report mean $\pm$ std for Dice, IoU, and NSD over all samples. The main table reports results under point prompting, while box-prompt results are provided in the supplementary material. The discussion below focuses on Dice for clarity.

RobustMedSAM achieves the best overall performance of \textbf{$0.719 \pm 0.193$}, outperforming SAM ($0.613 \pm 0.180$) by \textbf{+0.106} and RobustSAM ($0.632 \pm 0.202$) by \textbf{+0.087}. MedSAM shows severe degradation under point prompting (overall Dice $0.201 \pm 0.044$), while RobustMedSAM improves Dice by \textbf{+0.518}.


RobustMedSAM maintains strong performance across modalities, with the largest gains where RobustSAM lacks medical priors (e.g., PROMISE12: $+0.251$ over RobustSAM) and where SAM lacks robustness (e.g., AbdomenUS: $+0.203$ over SAM), confirming that robustness-oriented decoding alone is insufficient without medical-domain adaptation.

\begin{figure}
    \centering
    \includegraphics[width=0.95\linewidth]{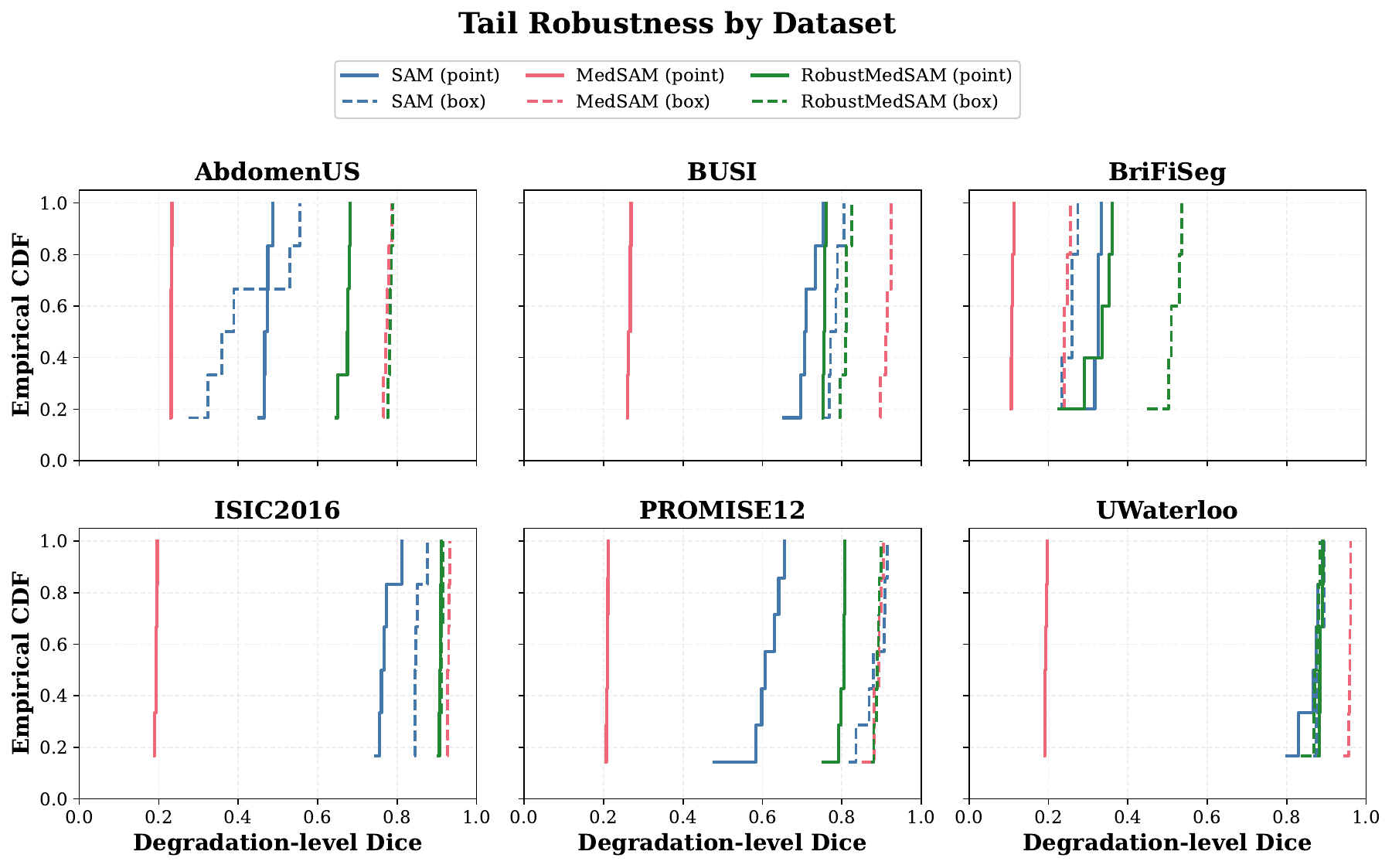}
    \caption{\textbf{Tail robustness under degradations.} Empirical CDF of degradation-level Dice across datasets. RobustMedSAM shows a consistent rightward shift under point prompts, indicating improved worst-case performance, while remaining comparable to MedSAM under box prompts.}
    \label{fig:tailcdf}
\end{figure}

Figure~\ref{fig:tailcdf} further illustrates the lower-tail behavior under degradations. Under point prompts, the empirical CDFs of RobustMedSAM consistently shift rightward relative to SAM and MedSAM, indicating improved Dice scores in the low-performance regime. Under box prompts (Suppl. Table~\ref{tab:bbox_main}), RobustMedSAM achieves overall Dice comparable to MedSAM ($0.804$ vs.\ $0.801$); however, MedSAM leads on IoU ($0.723$ vs.\ $0.706$), consistent with its box-centric fine-tuning. RobustMedSAM notably outperforms both SAM and MedSAM on BriFiSeg and modality-specific degradations (e.g., Poisson noise, Step Motion MRI), where robust decoding provides the largest benefit.

\begin{figure}
    \centering
    \includegraphics[width=0.9\linewidth]{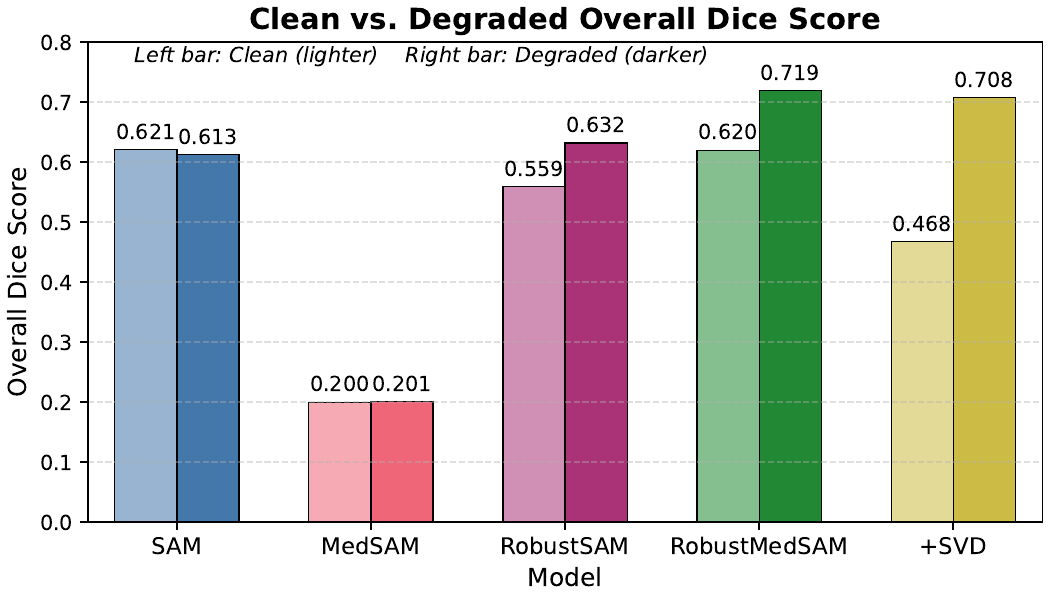}
    \caption{\textbf{Point Prompts: Clean vs. degraded comparison} (overall Dice ↑, point prompts). RobustMedSAM improves robustness on degraded inputs while maintaining clean performance comparable to SAM; +SVD exhibits a robustness–clean trade-off.}
    \label{fig:clean}
\end{figure}

To quantify the robustness–accuracy trade-off (i.e., whether robustness gains come at the cost of clean performance), we compare overall Dice on clean and degraded inputs in Figure~\ref{fig:clean}. RobustMedSAM is best on degraded images ($0.719$), surpassing SAM ($0.613$, \textbf{+0.106}) and RobustSAM ($0.632$, \textbf{+0.087}), while remaining on par with SAM on clean images ($0.620$ vs. $0.621$). In contrast, MedSAM collapses under point prompting ($\approx0.20$), and the +SVD variant shows a worse trade-off, reducing clean Dice ($0.468$) without improving degraded performance over RobustMedSAM ($0.708$ vs. $0.719$). Detailed results for each dataset are provided in the supplementary material (Table~\ref{tab:clean_vs_deg}).

\input{subsection/table_main}

\subsection{Per-Degradation Analysis}
\label{sec:degradation}
\begin{figure}
    \centering
    \includegraphics[width=0.9\linewidth]{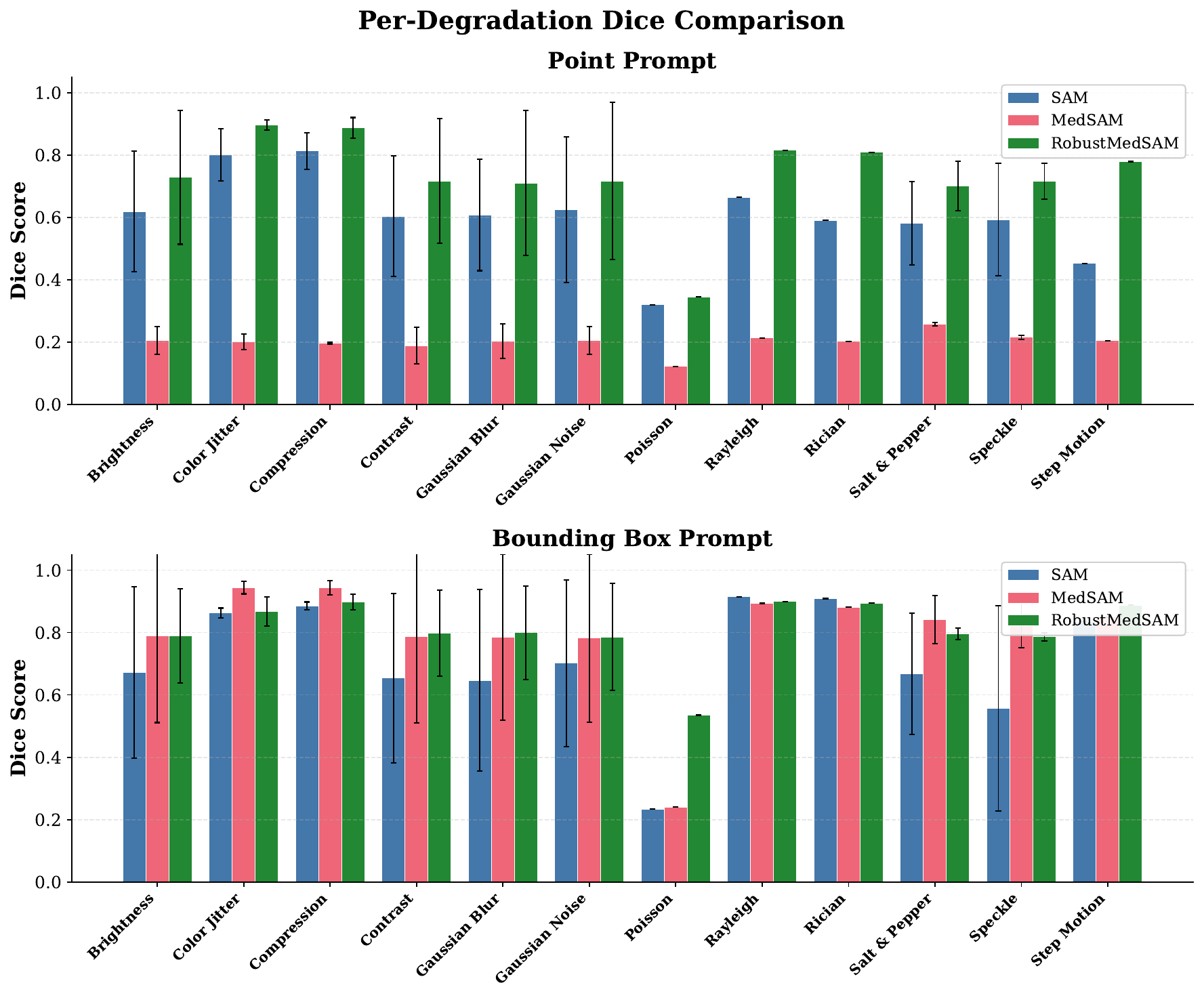}
    \caption{\textbf{Per-degradation analysis}. Dice comparison across degradation types under point and box prompts. RobustMedSAM shows the clearest gains under point prompting, while remaining competitive under box prompting.}
    \label{fig:degradation}
\end{figure}

Figure~\ref{fig:degradation} presents the performance breakdown across degradation types. Under point prompts, RobustMedSAM consistently outperforms both SAM and MedSAM across all degradations, with the largest gains observed for modality-specific corruptions commonly encountered in medical imaging. In particular, it shows pronounced improvements on step motion, Rician noise, and Rayleigh noise, while also improving robustness to brightness and contrast shifts, Gaussian blur, and speckle corruption. Under box prompts, RobustMedSAM remains competitive with MedSAM and shows clearer advantages on Poisson noise and step motion, whereas the gap is smaller on most other degradation types. Detailed results for each degradation are provided in the supplementary material (Tables~\ref{tab:degradation_dice}, \ref{tab:degradation_iou}, and \ref{tab:degradation_nsd}).

\subsection{Cross-Modality Analysis}
\label{sec:modality}
\begin{figure}
    \centering
    \includegraphics[width=0.95\linewidth]{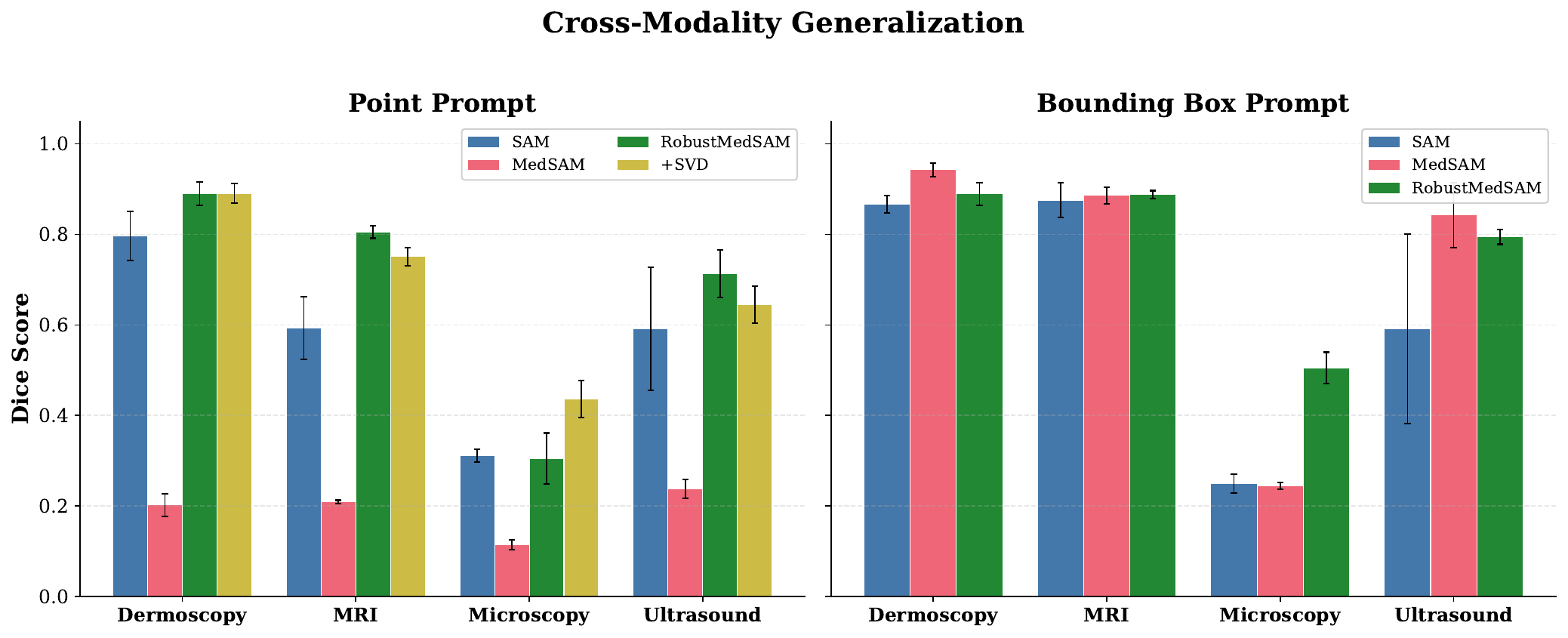}
    \caption{\textbf{Per-modality analysis}. RobustMedSAM shows the clearest gains under point prompts, especially on MRI and ultrasound, while remaining competitive under box prompts.}
    \label{fig:modality}
\end{figure}

Figure~\ref{fig:modality} summarizes cross-modality generalization. Under point prompts, RobustMedSAM achieves the strongest performance on dermoscopy, MRI, and ultrasound, with the largest gains over SAM and MedSAM observed on MRI and ultrasound. On microscopy, its performance remains limited, although still improving over MedSAM. Under box prompts, RobustMedSAM remains competitive on MRI and substantially improves microscopy, while MedSAM performs better on dermoscopy and ultrasound, indicating that robustness gains are more pronounced for point-based prompting than for box-based prompting.

\subsection{Out-of-Distribution Generalization}
\label{sec:ood}
We evaluate OOD generalization on six datasets with unseen modality--anatomy combinations (Table~\ref{tab:ood}) under the same single-image inference on degraded inputs. 
Overall, RobustMedSAM improves the average Dice from $0.455$ (SAM) to $\mathbf{0.493}$ (\textbf{+0.038}), outperforming RobustSAM ($0.477$). The largest gains appear on ISIC~2018 (\textbf{+0.083}) and PanDental (\textbf{+0.182}), where our method achieves the best performance. A small improvement is observed on DRIVE (0.224 vs.\ 0.209). Performance is comparable to SAM/RobustSAM on Kvasir, while gains are limited on thin-vessel datasets (CHASE\_DB1, DCA1), where severe class imbalance and topology sensitivity dominate.

\begin{table}
\centering
\caption{Out-of-distribution evaluation (Dice $\uparrow$, \textbf{point prompts}).}
\label{tab:ood}
\small
\resizebox{\linewidth}{!}{
\begin{tabular}{lccccc}
\toprule
Dataset & SAM & MedSAM & RobustSAM & Ours & +SVD \\
\midrule
CHASE\_DB1 & \cellcolor{datacellgreen}0.153 & 0.078 & 0.152 & 0.121 & 0.059 \\
DRIVE & 0.209 & 0.093 & 0.216 & \cellcolor{datacellgreen}0.224 & 0.141 \\
ISIC 2018 & 0.788 & 0.183 & 0.843 &\cellcolor{datacellgreen}0.871 & 0.870 \\
Kvasir & 0.770 & 0.195 & \cellcolor{datacellgreen}0.775 & 0.774 & 0.754 \\
PanDental & 0.638 & 0.106 & 0.668 & \cellcolor{datacellgreen}0.820 & 0.649 \\
DCA1 & 0.272 & 0.107 & \cellcolor{datacellgreen}0.313 & 0.297 & 0.217 \\
\midrule
\textbf{Overall} & 0.455 & 0.127 & 0.477 & \cellcolor{datacellgreen}0.493 & 0.427 \\
\bottomrule
\end{tabular}}
\end{table}

\subsection{Prompt Sensitivity Analysis}
\label{sec:prompt_sensitivity}
\begin{figure}[ht]
    \centering
    \includegraphics[width=0.9\linewidth]{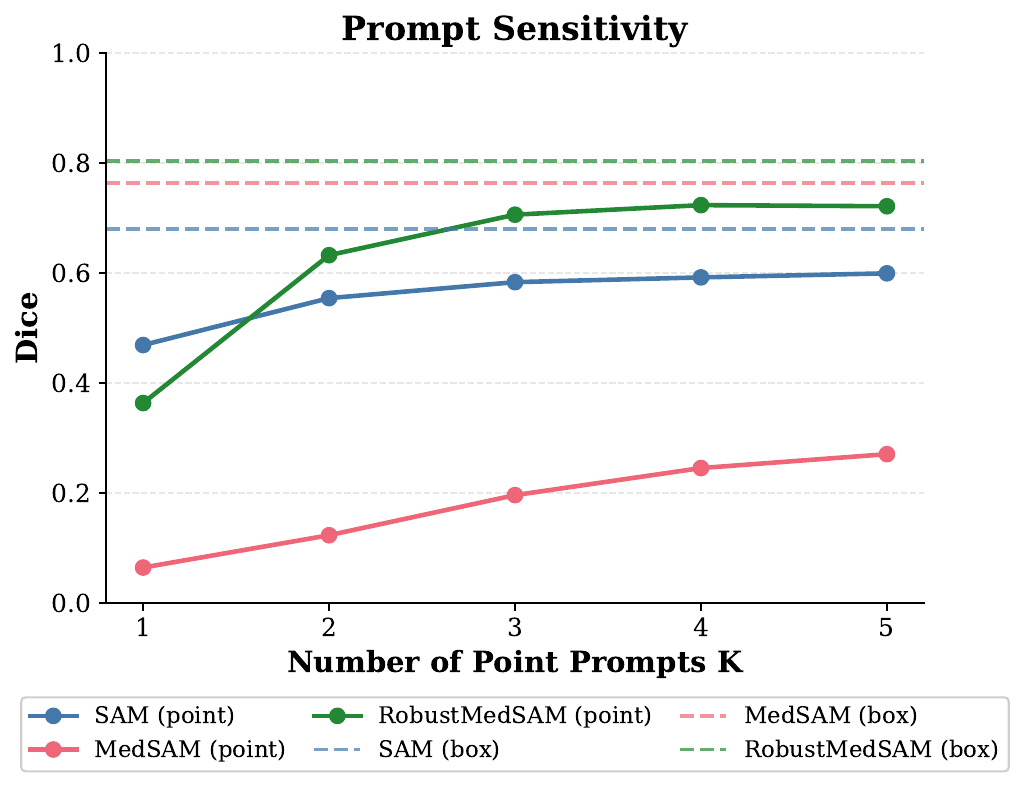}
    \caption{\textbf{Prompt sensitivity.} Dice versus the number of point prompts $K$. RobustMedSAM improves rapidly and saturates around $K=3$, while box-prompted models are shown as reference baselines.}
    \label{fig:prompt_sensitivity}
\end{figure}

Figure~\ref{fig:prompt_sensitivity} shows performance as a function of the number of point prompts $K$. RobustMedSAM improves rapidly as $K$ increases and largely saturates after $K=3$, whereas SAM exhibits smaller but steady gains and MedSAM remains substantially weaker under point prompting.

For reference, we also include box-prompted SAM, MedSAM, and RobustMedSAM. MedSAM remains highly sensitive to prompt type, performing poorly with points but strongly with boxes. In contrast, RobustMedSAM is more point-efficient: it surpasses point-prompted SAM for $K \geq 2$ and exceeds the box-prompted SAM baseline once $K \geq 3$. However, both MedSAM and RobustMedSAM still achieve higher performance under box prompting.

\subsection{Qualitative Results}
\label{sec:qualitative}
\begin{figure*}
    \centering
    \includegraphics[width=1\linewidth]{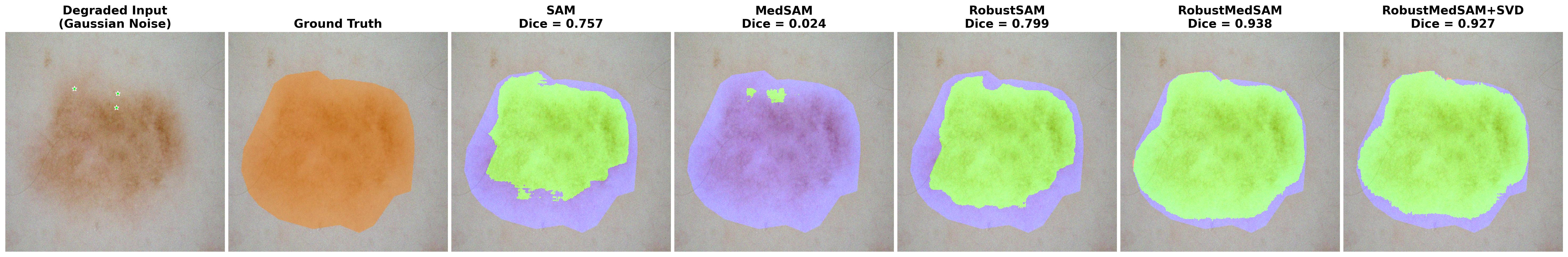}
    \hfill
    \includegraphics[width=1\linewidth]{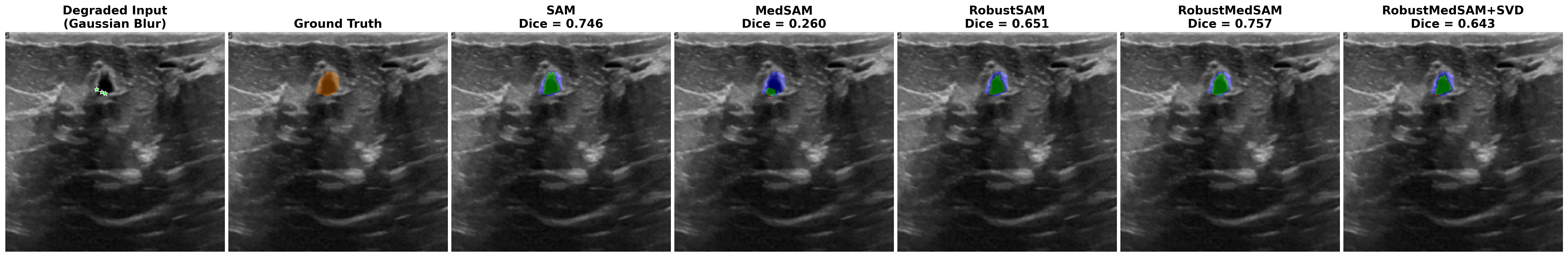}
    \\[4pt]
    \includegraphics[width=1\linewidth]{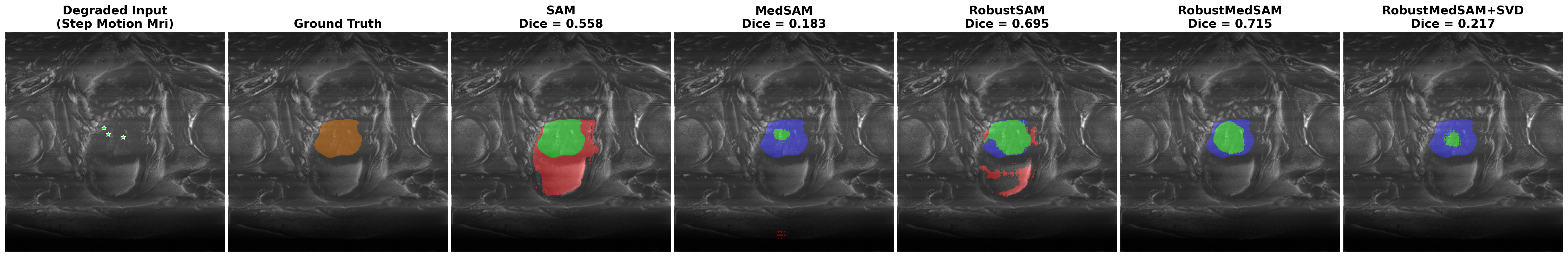}
    \hfill
    \includegraphics[width=1\linewidth]{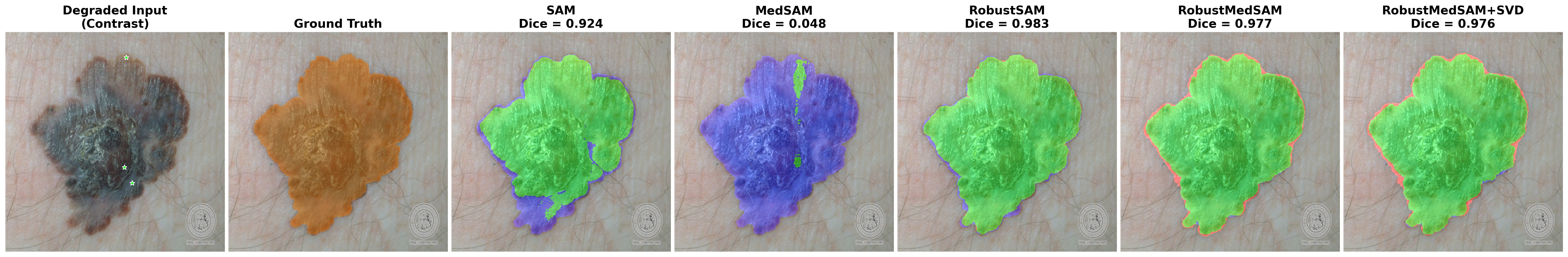}
    \caption{Qualitative segmentation results under degradations. Each panel shows (left to right): degraded input with \textbf{point prompts}, ground truth, SAM, MedSAM, RobustMedSAM, and RobustMedSAM+SVD predictions. Green = true positive, red = false positive, blue = false negative.}
    \label{fig:qualitative}
\end{figure*}

Figure~\ref{fig:qualitative} visualizes representative segmentations under degradations. Across modalities and corruption types, RobustMedSAM produces cleaner, more contiguous masks than SAM, which tends to fragment or under-segment under corruption, while MedSAM consistently fails under point prompting.


\subsection{Ablation: SVD Encoder Adaptation}
\label{sec:ablation}

Comparing RobustMedSAM and RobustMedSAM+SVD isolates the effect of SVD-based encoder adaptation. On degraded images, RobustMedSAM performs slightly better overall. However, +SVD improves several encoder-sensitive cases, notably Poisson noise and BriFiSeg. On clean images, RobustMedSAM matches SAM ($0.620$ vs.\ $0.621$, see Table~\ref{tab:clean_vs_deg}), whereas +SVD drops to $0.470$, indicating an unfavorable robustness--clean trade-off. We recommend freezing the encoder by default.

%% file: subsection/table_main.tex
\begin{table*}
\centering
\renewcommand{\arraystretch}{1.25}
\caption{Segmentation performance under various image degradations with \textbf{point prompts}. Values are reported as mean $\pm$ std across degradation types per dataset. Metrics: Dice $\uparrow$, IoU $\uparrow$, and boundary accuracy NSD $\uparrow$ ($\tau{=}2$\,px).}
\label{tab:main}
\small
\resizebox{\linewidth}{!}{
\begin{tabular}{lccccccccccccccc}
\toprule
& \multicolumn{5}{c}{Dice $\uparrow$} & \multicolumn{5}{c}{IoU $\uparrow$} & \multicolumn{5}{c}{NSD $\uparrow$ ($\tau{=}2$\,px)} \\
\cmidrule(lr){2-6}\cmidrule(lr){7-11}\cmidrule(lr){12-16}
Dataset &
\makecell[c]{SAM} &
\makecell[c]{MedSAM} &
\makecell[c]{Robust\\SAM} &
\makecell[c]{Robust\\MedSAM} &
\makecell[c]{Robust\\MedSAM\\+SVD} &
\makecell[c]{SAM} &
\makecell[c]{MedSAM} &
\makecell[c]{Robust\\SAM} &
\makecell[c]{Robust\\MedSAM} &
\makecell[c]{Robust\\MedSAM\\+SVD} &
\makecell[c]{SAM} &
\makecell[c]{MedSAM} &
\makecell[c]{Robust\\SAM} &
\makecell[c]{Robust\\MedSAM} &
\makecell[c]{Robust\\MedSAM\\+SVD} \\
\midrule

ISIC 2016 &
\makecell{$0.752$ \\ {\small ($\pm 0.033$)}} &
\makecell{$0.202$ \\ {\small ($\pm 0.013$)}} &
\makecell{$0.836$ \\ {\small ($\pm 0.019$)}} &
\cellcolor{datacellgreen}\makecell{$0.911$ \\ {\small ($\pm 0.003$)}} &
\makecell{$0.910$ \\ {\small ($\pm 0.006$)}} &
\makecell{$0.640$ \\ {\small ($\pm 0.037$)}} &
\makecell{$0.120$ \\ {\small ($\pm 0.009$)}} &
\makecell{$0.742$ \\ {\small ($\pm 0.022$)}} &
\cellcolor{datacellgreen}\makecell{$0.844$ \\ {\small ($\pm 0.005$)}} &
\makecell{$0.842$ \\ {\small ($\pm 0.006$)}} &
\makecell{$0.088$ \\ {\small ($\pm 0.024$)}} &
\makecell{$0.017$ \\ {\small ($\pm 0.001$)}} &
\makecell{$0.135$ \\ {\small ($\pm 0.021$)}} &
\cellcolor{datacellgreen}\makecell{\textbf{$0.247$} \\ {\small \textbf{($\pm 0.009$)}}} &
\makecell{$0.232$ \\ {\small ($\pm 0.007$)}} \\

BUSI &
\makecell{$0.718$ \\ {\small ($\pm 0.036$)}} &
\makecell{$0.249$ \\ {\small ($\pm 0.016$)}} &
\makecell{$0.686$ \\ {\small ($\pm 0.061$)}} &
\cellcolor{datacellgreen}\makecell{$0.760$ \\ {\small ($\pm 0.017$)}} &
\makecell{$0.678$ \\ {\small ($\pm 0.032$)}} &
\makecell{$0.609$ \\ {\small ($\pm 0.039$)}} &
\makecell{$0.152$ \\ {\small ($\pm 0.010$)}} &
\makecell{$0.578$ \\ {\small ($\pm 0.061$)}} &
\cellcolor{datacellgreen}\makecell{$0.642$ \\ {\small ($\pm 0.018$)}} &
\makecell{$0.561$ \\ {\small ($\pm 0.028$)}} &
\makecell{$0.226$ \\ {\small ($\pm 0.020$)}} &
\makecell{$0.044$ \\ {\small ($\pm 0.001$)}} &
\makecell{$0.223$ \\ {\small ($\pm 0.030$)}} &
\cellcolor{datacellgreen}\makecell{\textbf{$0.228$} \\ {\small \textbf{($\pm 0.004$)}}} &
\makecell{$0.206$ \\ {\small ($\pm 0.012$)}} \\

PROMISE12 &
\makecell{$0.593$ \\ {\small ($\pm 0.069$)}} &
\makecell{$0.209$ \\ {\small ($\pm 0.004$)}} &
\makecell{$0.554$ \\ {\small ($\pm 0.044$)}} &
\cellcolor{datacellgreen}\makecell{$0.805$ \\ {\small ($\pm 0.014$)}} &
\makecell{$0.751$ \\ {\small ($\pm 0.020$)}} &
\makecell{$0.481$ \\ {\small ($\pm 0.069$)}} &
\makecell{$0.123$ \\ {\small ($\pm 0.003$)}} &
\makecell{$0.426$ \\ {\small ($\pm 0.042$)}} &
\cellcolor{datacellgreen}\makecell{$0.687$ \\ {\small ($\pm 0.019$)}} &
\makecell{$0.620$ \\ {\small ($\pm 0.026$)}} &
\makecell{$0.176$ \\ {\small ($\pm 0.045$)}} &
\makecell{$0.028$ \\ {\small ($\pm 0.001$)}} &
\makecell{$0.122$ \\ {\small ($\pm 0.023$)}} &
\cellcolor{datacellgreen}\makecell{\textbf{$0.199$} \\ {\small \textbf{($\pm 0.018$)}}} &
\makecell{$0.177$ \\ {\small ($\pm 0.012$)}} \\

UWaterloo Skin Cancer &
\makecell{$0.841$ \\ {\small ($\pm 0.027$)}} &
\makecell{$0.202$ \\ {\small ($\pm 0.034$)}} &
\cellcolor{datacellgreen}\makecell{$0.914$ \\ {\small ($\pm 0.007$)}} &
\makecell{$0.869$ \\ {\small ($\pm 0.019$)}} &
\makecell{$0.872$ \\ {\small ($\pm 0.010$)}} &
\makecell{$0.757$ \\ {\small ($\pm 0.033$)}} &
\makecell{$0.120$ \\ {\small ($\pm 0.023$)}} &
\cellcolor{datacellgreen}\makecell{$0.850$ \\ {\small ($\pm 0.009$)}} &
\makecell{$0.782$ \\ {\small ($\pm 0.022$)}} &
\makecell{$0.785$ \\ {\small ($\pm 0.011$)}} &
\makecell{$0.351$ \\ {\small ($\pm 0.038$)}} &
\makecell{$0.046$ \\ {\small ($\pm 0.008$)}} &
\cellcolor{datacellgreen}\makecell{\textbf{$0.444$} \\ {\small \textbf{($\pm 0.038$)}}} &
\makecell{$0.295$ \\ {\small ($\pm 0.011$)}} &
\makecell{$0.270$ \\ {\small ($\pm 0.013$)}} \\

BriFiSeg &
\makecell{$0.311$ \\ {\small ($\pm 0.014$)}} &
\makecell{$0.114$ \\ {\small ($\pm 0.011$)}} &
\makecell{$0.329$ \\ {\small ($\pm 0.022$)}} &
\makecell{$0.305$ \\ {\small ($\pm 0.056$)}} &
\cellcolor{datacellgreen}\makecell{$0.436$ \\ {\small ($\pm 0.041$)}} &
\makecell{$0.188$ \\ {\small ($\pm 0.010$)}} &
\makecell{$0.062$ \\ {\small ($\pm 0.006$)}} &
\makecell{$0.201$ \\ {\small ($\pm 0.016$)}} &
\makecell{$0.194$ \\ {\small ($\pm 0.041$)}} &
\cellcolor{datacellgreen}\makecell{$0.294$ \\ {\small ($\pm 0.034$)}} &
\makecell{$0.108$ \\ {\small ($\pm 0.004$)}} &
\makecell{$0.049$ \\ {\small ($\pm 0.001$)}} &
\makecell{$0.104$ \\ {\small ($\pm 0.002$)}} &
\makecell{$0.129$ \\ {\small ($\pm 0.015$)}} &
\cellcolor{datacellgreen}\makecell{\textbf{$0.183$} \\ {\small \textbf{($\pm 0.009$)}}} \\

AbdomenUS &
\makecell{$0.463$ \\ {\small ($\pm 0.019$)}} &
\makecell{$0.227$ \\ {\small ($\pm 0.020$)}} &
\makecell{$0.473$ \\ {\small ($\pm 0.007$)}} &
\cellcolor{datacellgreen}\makecell{$0.666$ \\ {\small ($\pm 0.023$)}} &
\makecell{$0.612$ \\ {\small ($\pm 0.012$)}} &
\makecell{$0.331$ \\ {\small ($\pm 0.014$)}} &
\makecell{$0.139$ \\ {\small ($\pm 0.015$)}} &
\makecell{$0.339$ \\ {\small ($\pm 0.007$)}} &
\cellcolor{datacellgreen}\makecell{$0.523$ \\ {\small ($\pm 0.024$)}} &
\makecell{$0.472$ \\ {\small ($\pm 0.012$)}} &
\makecell{$0.133$ \\ {\small ($\pm 0.034$)}} &
\makecell{$0.039$ \\ {\small ($\pm 0.001$)}} &
\cellcolor{datacellgreen}\makecell{\textbf{$0.175$} \\ {\small \textbf{($\pm 0.036$)}}} &
\makecell{$0.105$ \\ {\small ($\pm 0.007$)}} &
\makecell{$0.086$ \\ {\small ($\pm 0.007$)}} \\

\midrule
\textbf{Overall} &
\makecell{$0.613$ \\ {\small ($\pm 0.180$)}} &
\makecell{$0.201$ \\ {\small ($\pm 0.044$)}} &
\makecell{$0.632$ \\ {\small ($\pm 0.202$)}} &
\cellcolor{datacellgreen}\makecell{$0.719$ \\ {\small ($\pm 0.193$)}} &
\makecell{$0.710$ \\ {\small ($\pm 0.157$)}} &
\makecell{$0.501$ \\ {\small ($\pm 0.192$)}} &
\makecell{$0.119$ \\ {\small ($\pm 0.029$)}} &
\makecell{$0.523$ \\ {\small ($\pm 0.225$)}} &
\cellcolor{datacellgreen}\makecell{$0.612$ \\ {\small ($\pm 0.205$)}} &
\makecell{$0.596$ \\ {\small ($\pm 0.181$)}} &
\makecell{$0.180$ \\ {\small ($\pm 0.094$)}} &
\makecell{$0.037$ \\ {\small ($\pm 0.012$)}} &
\makecell{$0.200$ \\ {\small ($\pm 0.120$)}} &
\cellcolor{datacellgreen}\makecell{\textbf{$0.201$} \\ {\small \textbf{($\pm 0.067$)}}} &
\makecell{$0.192$ \\ {\small ($\pm 0.058$)}} \\
\bottomrule
\end{tabular}}
\end{table*}

%% file: subsection/5-conclusion.tex
\section{Discussion}
\label{sec:discussion}
\noindent\textbf{Why does fusion help?}
RobustMedSAM exploits a separation between \emph{what} to represent and \emph{how} to decode. MedSAM's encoder provides anatomy-aware features lacking in generic SAM, while RobustSAM's decoder is trained to remain stable under corruptions. The shared SAM ViT-B interface enables checkpoint fusion, but the gains are not a trivial initialization effect: RobustSAM is competitive on some datasets but degrades on others (e.g., PROMISE12), whereas the fused model improves consistently and shifts the worst-case Dice distribution rightward.

\noindent\textbf{Robustness without sacrificing clean accuracy.}
A concern is that robustness harms clean performance. Figure~\ref{fig:clean} shows that freezing the MedSAM encoder avoids this: RobustMedSAM matches SAM on clean inputs while improving degraded Dice by +0.106, suggesting robustness can be addressed mainly in the decoder.

\noindent\textbf{Encoder adaptation trade-off.}
The +SVD variant improves a few encoder-sensitive cases (e.g., Poisson noise) but reduces clean accuracy, consistent with shifting features toward corrupted inputs. We therefore recommend freezing the encoder by default and using adaptation only when images are persistently degraded.

\noindent\textbf{Prompt mismatch and clinical usability.}
MedSAM’s weak point-prompt performance largely reflects its box-centric fine-tuning. To control for this mismatch, we include box-prompted baselines and evaluate all point-prompt methods under a shared $K{=}3$ protocol. In practice, interactive segmentation cannot rely on a fixed prompt type; RobustMedSAM is less prompt-sensitive, remains competitive under box prompting, and saturates with few points.

\noindent\textbf{Limitations.}
Several limitations should be noted. First, gains are limited for topology-sensitive thin vessels and microscopy, where fine-grained encoder features are critical and decoder-side robustness alone cannot compensate,  suggesting that module-wise fusion is most effective when the required capabilities map cleanly onto distinct architectural components. Second, MedSegBench uses synthetic corruptions; validation on real clinical artifacts (e.g., patient motion during acquisition, hardware-specific noise profiles) remains necessary to confirm practical utility. Finally, only RobustMedSAM receives degradation-aware training; a corruption-trained MedSAM baseline would more cleanly isolate the contribution of module-wise fusion from the benefit of degradation-augmented training, and constitutes important future work.

\section{Conclusion}
\label{sec:conclusion}
Our experiments demonstrate that module-wise checkpoint fusion, initializing different architectural components from models trained for complementary tasks, effectively composes medical-domain adaptation with corruption robustness for medical image segmentation. By fusing MedSAM’s medical-domain encoder with RobustSAM’s robust decoder and training only the decoder, the method preserves anatomical feature quality while improving robustness under image corruption. Across 35 datasets, six modalities, and 12 degradation types, RobustMedSAM improves degraded-image Dice from 0.613 to 0.719 over SAM and shows improvements on most out-of-distribution datasets, with limited gains on topology-sensitive cases. These results suggest that module-wise fusion of complementary pretrained models is a practical direction for robust medical segmentation under realistic imaging conditions.

\section*{Acknowledgment}
This work used the Hive cluster, which is supported by the National Science Foundation under grant number 1828187.  This research was supported in part through research cyberinfrastructure resources and services provided by the Partnership for an Advanced Computing Environment (PACE) at the Georgia Institute of Technology, Atlanta, Georgia, USA. RRID:SCR\_027619. We also gratefully acknowledge funding and fellowships that contributed to this work, including a Wallace H. Coulter Distinguished Faculty Fellowship, a Petit Institute Faculty Fellowship, and research funding from Amazon and Microsoft Research awarded to Professor May D. Wang.

%% file: subsection/6-supplementary.tex
\clearpage
\setcounter{page}{1}
\maketitlesupplementary

\setcounter{section}{0}
\setcounter{figure}{0}
\setcounter{table}{0}
\setcounter{equation}{0}

\renewcommand{\thesection}{S\arabic{section}}
\renewcommand{\thefigure}{S\arabic{figure}}
\renewcommand{\thetable}{S\arabic{table}}
\renewcommand{\theequation}{S\arabic{equation}}

\section{Evaluation Metrics}
\label{sec:metrics}
Let $P$ denote the predicted mask and $G$ the ground-truth mask.
\paragraph{Dice Coefficient}
\begin{equation}
\mathrm{Dice}(P,G) = \frac{2|P \cap G|}{|P| + |G|}.
\end{equation}

\paragraph{Intersection over Union (IoU)}
\begin{equation}
\mathrm{IoU}(P,G) = \frac{|P \cap G|}{|P \cup G|}.
\end{equation}
\begin{figure}
    \centering
    \includegraphics[width=\linewidth]{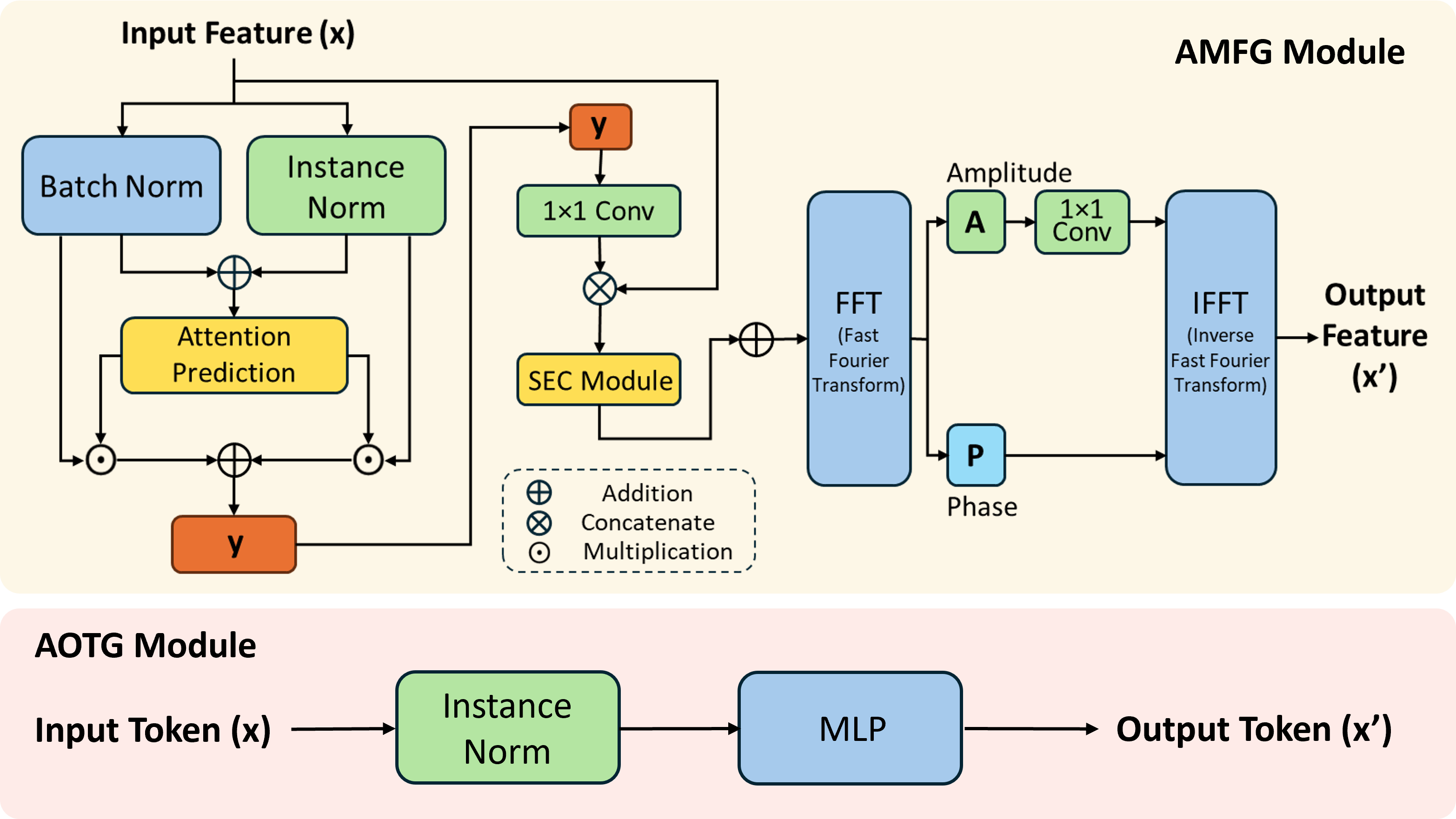}
    \caption{Detailed architecture of Anti-degradation Mask Feature Generation and Anti-degradation Output Token Generation modules.}
    \label{fig:module}
\end{figure}
\paragraph{Normalized Surface Distance (NSD)}
Let $\partial P$ and $\partial G$ denote the surfaces of $P$ and $G$. 
Given tolerance $\tau$, NSD is defined as
\begin{equation}
\mathrm{NSD}(P,G)=
\frac{
\begin{aligned}
&|\{x\in\partial P:d(x,\partial G)\le\tau\}| \\
+\, &|\{y\in\partial G:d(y,\partial P)\le\tau\}|
\end{aligned}
}{
|\partial P|+|\partial G|}
\end{equation}
where $d(\cdot,\partial G)$ denotes the shortest Euclidean distance to surface $\partial G$.

\section{Loss Weight Selection}
We set loss weights to balance term magnitude and functional role: $\alpha{=}20$ compensates for severe foreground--background imbalance and emphasizes hard pixels, while $\beta{=}1$ keeps overlap optimization on a stable scale. Since decoder-side MSE regularizers are typically low-magnitude, we set $\lambda_1{=}100$ as the primary constraint for degradation-invariant decoding and $\lambda_2{=}2$ as a token-level stabilizer to avoid suppressing $\mathcal{L}_{\text{seg}}$.

\section{Architecture Details}
\label{sec:arch_detail}
\subsection{Anti-degradation Mask Feature Generation}
We adopt the Anti-degradation Mask Feature Generation module proposed in~\cite{chen2024robustsam} to enhance feature robustness under image degradation. Given an input feature map $F$, the module first processes features using two parallel normalization
branches. Instance Normalization (IN) is used to reduce degradation-related style variations, while Batch Normalization (BN) preserves structural details that may be weakened by IN. The outputs of the two branches are fused and refined using an attention mechanism to adaptively balance their contributions.
\input{subsection/table_clear_vs_degraded}
The fused features are concatenated with the original input along the channel dimension and further refined by a channel attention module. To suppress degradation patterns, the features are transformed into the frequency domain using the Fourier transform. The amplitude component captures degradation-related style information and is filtered using a $1\times1$ convolution, while the phase component preserves structural information. The refined features are then transformed back to the spatial domain via the inverse Fourier transform.

\subsection{Anti-degradation Output Token Generation}
To enhance robustness against image degradation, we adopt
an Anti-Degradation Output Token Generation module proposed in~\cite{chen2024robustsam} to refine the Robust Output Token per mask ($T_{RO}$). Compared with mask features, $T_{RO}$ mainly encodes classification boundary information and contains limited texture details. Therefore, a lightweight refinement module is sufficient to suppress degradation-sensitive information.
\input{subsection/table_dice_degradation}
\input{subsection/table_iou_degradation}
Specifically, the token is processed through several layers of Instance Normalization followed by a single MLP layer. Instance Normalization reduces degradation-related style
variations, while the MLP further refines the token representation with minimal computational overhead.

The refined token is finally combined with the Robust Mask Feature to produce the final segmentation mask.
\section{Additional Experimental Results}
\subsection{Performance on Clear vs Degraded Images}
Table~\ref{tab:clean_vs_deg} reports segmentation performance under clean and degraded imaging conditions. The vanilla models (SAM and MedSAM) show negligible or negative changes overall, indicating limited robustness to degraded inputs. In contrast, the robust variants, particularly the SVD-based method, consistently benefit from the clear–degraded pair protocol and achieve substantial improvements across datasets, with the largest overall gain of +0.240.
\begin{figure*}
    \centering
    \includegraphics[width=\linewidth]{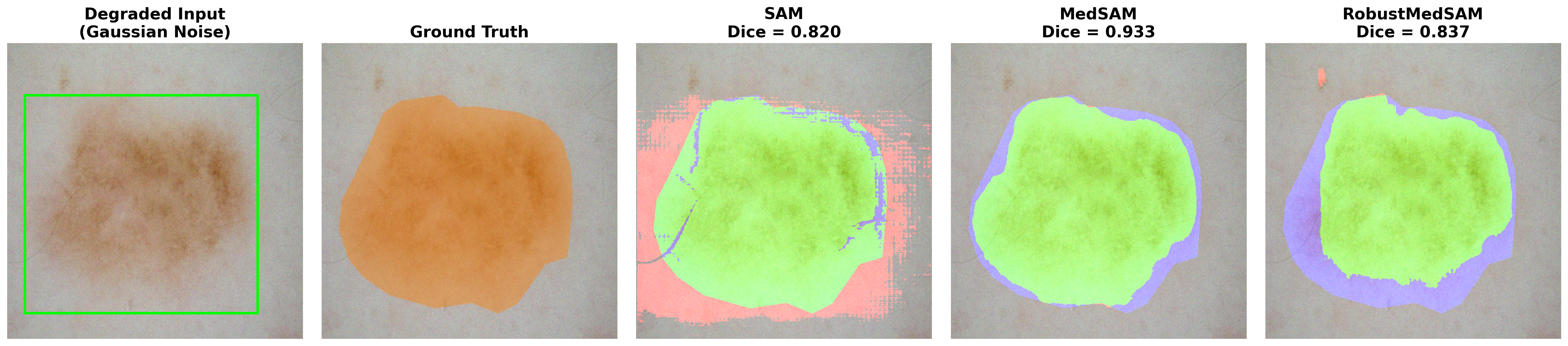}
    \hfill
    \includegraphics[width=1\linewidth]{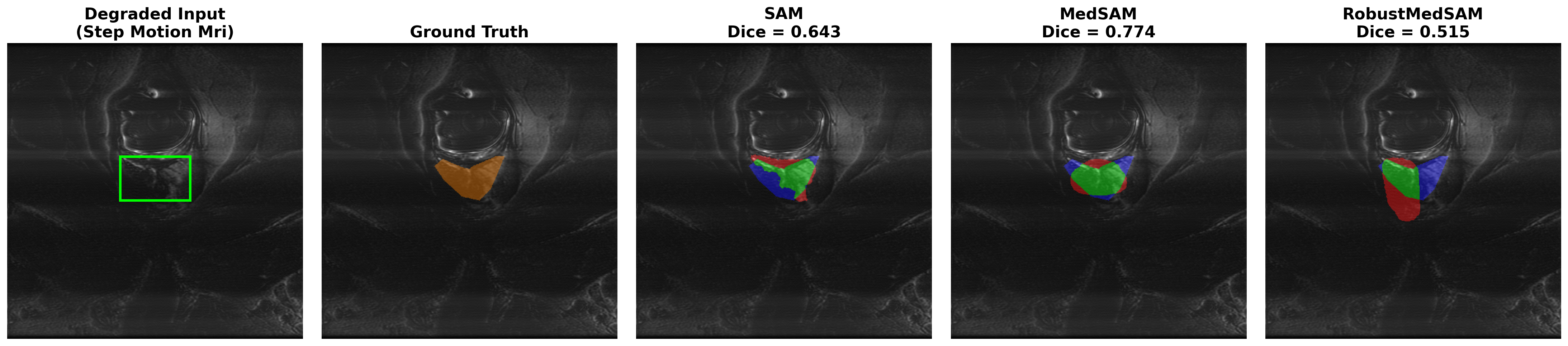}
    \hfill
    \includegraphics[width=1\linewidth]{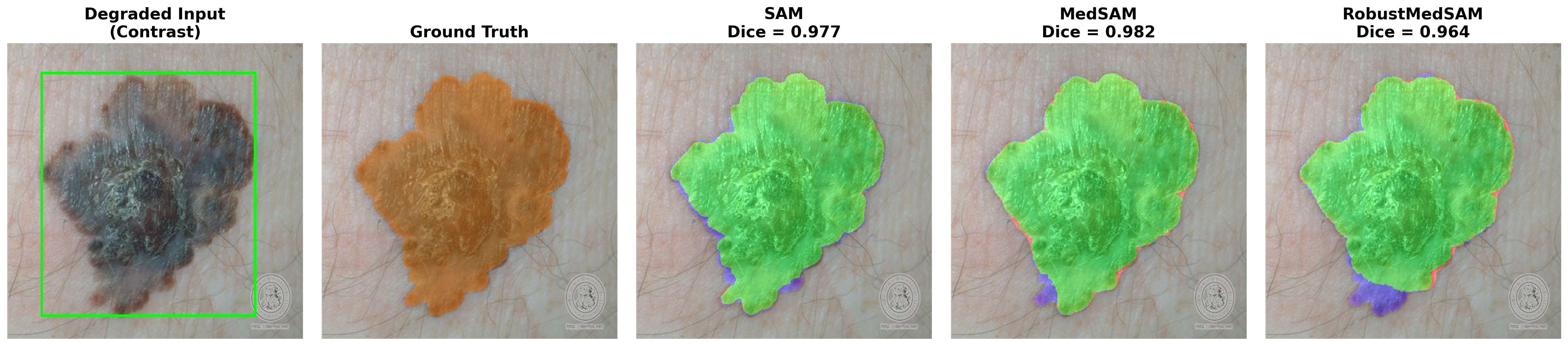}
    \caption{\textbf{Bounding Box:} Qualitative segmentation results under degradation. Each panel shows (left to right): degraded input with "bounding box prompts, ground truth, SAM, MedSAM, RobustMedSAM predictions. Green = true positive, red = false positive, blue = false negative.}
    \label{fig:qualitativez}
\end{figure*}
\subsection{Detailed Per-Degradation Analysis}
Tables~\ref{tab:degradation_dice}~\ref{tab:degradation_iou}~\ref{tab:degradation_nsd} presents the Dice, IoU and NSD of all models under different degradation types. Overall, the robust variants consistently outperform the vanilla SAM and MedSAM across most degradations, indicating improved robustness to diverse image corruptions. In particular, the SVD-enhanced variant shows notable advantages for certain noise-related degradations (e.g., compression and Poisson noise), while RobustMedSAM achieves strong and stable performance across most other degradation types.
\input{subsection/table_nsd_degradation}

\begin{table*}
\centering
\renewcommand{\arraystretch}{1.25}
\caption{\textbf{Bounding box prompt} segmentation performance under various image degradations. Values are reported as mean $\pm$ std across degradation types per dataset.}
\label{tab:bbox_main}
\small
\begin{tabular}{lcccccc}
\toprule
& \multicolumn{3}{c}{Dice $\uparrow$} & \multicolumn{3}{c}{IoU $\uparrow$} \\
\cmidrule(lr){2-4}\cmidrule(lr){5-7}
Dataset & \makecell[c]{SAM} & \makecell[c]{MedSAM} & \makecell[c]{Robust\\MedSAM} & \makecell[c]{SAM} & \makecell[c]{MedSAM} & \makecell[c]{Robust\\MedSAM} \\
\midrule
ISIC 2016 &
\makecell{$0.851$ \\ {\small ($\pm 0.012$)}} &
\cellcolor{datacellgreen}\makecell{$0.929$ \\ {\small ($\pm 0.003$)}} &
\makecell{$0.910$ \\ {\small ($\pm 0.006$)}} &
\makecell{$0.747$ \\ {\small ($\pm 0.018$)}} &
\cellcolor{datacellgreen}\makecell{$0.869$ \\ {\small ($\pm 0.004$)}} &
\makecell{$0.844$ \\ {\small ($\pm 0.009$)}} \\
BUSI &
\makecell{$0.778$ \\ {\small ($\pm 0.019$)}} &
\cellcolor{datacellgreen}\makecell{$0.911$ \\ {\small ($\pm 0.013$)}} &
\makecell{$0.807$ \\ {\small ($\pm 0.012$)}} &
\makecell{$0.683$ \\ {\small ($\pm 0.020$)}} &
\cellcolor{datacellgreen}\makecell{$0.840$ \\ {\small ($\pm 0.020$)}} &
\makecell{$0.701$ \\ {\small ($\pm 0.012$)}} \\
PROMISE12 &
\makecell{$0.876$ \\ {\small ($\pm 0.038$)}} &
\makecell{$0.886$ \\ {\small ($\pm 0.018$)}} &
\cellcolor{datacellgreen}\makecell{$0.888$ \\ {\small ($\pm 0.009$)}} &
\makecell{$0.793$ \\ {\small ($\pm 0.048$)}} &
\makecell{$0.803$ \\ {\small ($\pm 0.028$)}} &
\cellcolor{datacellgreen}\makecell{$0.804$ \\ {\small ($\pm 0.012$)}} \\
UWaterloo Skin Cancer &
\makecell{$0.881$ \\ {\small ($\pm 0.011$)}} &
\cellcolor{datacellgreen}\makecell{$0.956$ \\ {\small ($\pm 0.007$)}} &
\makecell{$0.869$ \\ {\small ($\pm 0.017$)}} &
\makecell{$0.797$ \\ {\small ($\pm 0.017$)}} &
\cellcolor{datacellgreen}\makecell{$0.917$ \\ {\small ($\pm 0.012$)}} &
\makecell{$0.791$ \\ {\small ($\pm 0.019$)}} \\
BriFiSeg &
\makecell{$0.250$ \\ {\small ($\pm 0.021$)}} &
\makecell{$0.244$ \\ {\small ($\pm 0.007$)}} &
\cellcolor{datacellgreen}\makecell{$0.505$ \\ {\small ($\pm 0.035$)}} &
\makecell{$0.146$ \\ {\small ($\pm 0.013$)}} &
\makecell{$0.142$ \\ {\small ($\pm 0.005$)}} &
\cellcolor{datacellgreen}\makecell{$0.348$ \\ {\small ($\pm 0.031$)}} \\
AbdomenUS &
\makecell{$0.405$ \\ {\small ($\pm 0.113$)}} &
\makecell{$0.774$ \\ {\small ($\pm 0.008$)}} &
\cellcolor{datacellgreen}\makecell{$0.782$ \\ {\small ($\pm 0.006$)}} &
\makecell{$0.307$ \\ {\small ($\pm 0.096$)}} &
\makecell{$0.654$ \\ {\small ($\pm 0.010$)}} &
\cellcolor{datacellgreen}\makecell{$0.673$ \\ {\small ($\pm 0.006$)}} \\
\midrule
\textbf{Overall} &
\makecell{$0.691$ \\ {\small ($\pm 0.251$)}} &
\makecell{$0.801$ \\ {\small ($\pm 0.234$)}} &
\cellcolor{datacellgreen}\makecell{$0.804$ \\ {\small ($\pm 0.131$)}} &
\makecell{$0.597$ \\ {\small ($\pm 0.255$)}} &
\cellcolor{datacellgreen}\makecell{$0.723$ \\ {\small ($\pm 0.251$)}} &
\makecell{$0.706$ \\ {\small ($\pm 0.159$)}} \\
\bottomrule
\end{tabular}
\end{table*}

\begin{table*}
\centering
\renewcommand{\arraystretch}{1.25}
\caption{Per-degradation analysis \textbf{(bounding box prompt)}. Values are mean $\pm$ std across datasets.}
\label{tab:bbox_degradation}
\small
\begin{tabular}{lcccccc}
\toprule
& \multicolumn{3}{c}{Dice $\uparrow$} & \multicolumn{3}{c}{IoU $\uparrow$} \\
\cmidrule(lr){2-4}\cmidrule(lr){5-7}
\makecell[l]{Degradation} & \makecell[c]{SAM} & \makecell[c]{MedSAM} & \makecell[c]{Robust\\MedSAM} & \makecell[c]{SAM} & \makecell[c]{MedSAM} & \makecell[c]{Robust\\MedSAM} \\
\midrule
Brightness &
\makecell{$0.671$ \\ {\small $\pm 0.275$}} &
\makecell{$0.788$ \\ {\small $\pm 0.277$}} &
\cellcolor{datacellgreen}\makecell{$0.790$ \\ {\small $\pm 0.151$}} &
\makecell{$0.575$ \\ {\small $\pm 0.281$}} &
\cellcolor{datacellgreen}\makecell{$0.713$ \\ {\small $\pm 0.297$}} &
\makecell{$0.689$ \\ {\small $\pm 0.183$}} \\
Color Jitter &
\makecell{$0.862$ \\ {\small $\pm 0.016$}} &
\cellcolor{datacellgreen}\makecell{$0.944$ \\ {\small $\pm 0.020$}} &
\makecell{$0.868$ \\ {\small $\pm 0.046$}} &
\makecell{$0.766$ \\ {\small $\pm 0.029$}} &
\cellcolor{datacellgreen}\makecell{$0.895$ \\ {\small $\pm 0.034$}} &
\makecell{$0.795$ \\ {\small $\pm 0.052$}} \\
Compression &
\makecell{$0.885$ \\ {\small $\pm 0.013$}} &
\cellcolor{datacellgreen}\makecell{$0.943$ \\ {\small $\pm 0.022$}} &
\makecell{$0.898$ \\ {\small $\pm 0.026$}} &
\makecell{$0.799$ \\ {\small $\pm 0.023$}} &
\cellcolor{datacellgreen}\makecell{$0.894$ \\ {\small $\pm 0.039$}} &
\makecell{$0.829$ \\ {\small $\pm 0.035$}} \\
Contrast &
\makecell{$0.654$ \\ {\small $\pm 0.271$}} &
\makecell{$0.786$ \\ {\small $\pm 0.276$}} &
\cellcolor{datacellgreen}\makecell{$0.798$ \\ {\small $\pm 0.138$}} &
\makecell{$0.557$ \\ {\small $\pm 0.275$}} &
\cellcolor{datacellgreen}\makecell{$0.710$ \\ {\small $\pm 0.296$}} &
\makecell{$0.696$ \\ {\small $\pm 0.171$}} \\
Gaussian Blur &
\makecell{$0.647$ \\ {\small $\pm 0.292$}} &
\makecell{$0.785$ \\ {\small $\pm 0.266$}} &
\cellcolor{datacellgreen}\makecell{$0.800$ \\ {\small $\pm 0.150$}} &
\makecell{$0.552$ \\ {\small $\pm 0.290$}} &
\cellcolor{datacellgreen}\makecell{$0.705$ \\ {\small $\pm 0.284$}} &
\makecell{$0.701$ \\ {\small $\pm 0.183$}} \\
Gaussian Noise &
\makecell{$0.702$ \\ {\small $\pm 0.267$}} &
\makecell{$0.782$ \\ {\small $\pm 0.270$}} &
\cellcolor{datacellgreen}\makecell{$0.785$ \\ {\small $\pm 0.172$}} &
\makecell{$0.607$ \\ {\small $\pm 0.276$}} &
\cellcolor{datacellgreen}\makecell{$0.702$ \\ {\small $\pm 0.288$}} &
\makecell{$0.686$ \\ {\small $\pm 0.199$}} \\
Poisson &
\makecell{$0.233$ \\ {\small $\pm 0.101$}} &
\makecell{$0.240$ \\ {\small $\pm 0.092$}} &
\cellcolor{datacellgreen}\makecell{$0.535$ \\ {\small $\pm 0.123$}} &
\makecell{$0.136$ \\ {\small $\pm 0.064$}} &
\makecell{$0.139$ \\ {\small $\pm 0.059$}} &
\cellcolor{datacellgreen}\makecell{$0.375$ \\ {\small $\pm 0.113$}} \\
Rayleigh &
\cellcolor{datacellgreen}\makecell{$0.915$ \\ {\small $\pm 0.042$}} &
\makecell{$0.894$ \\ {\small $\pm 0.077$}} &
\makecell{$0.899$ \\ {\small $\pm 0.061$}} &
\cellcolor{datacellgreen}\makecell{$0.846$ \\ {\small $\pm 0.068$}} &
\makecell{$0.816$ \\ {\small $\pm 0.115$}} &
\makecell{$0.822$ \\ {\small $\pm 0.084$}} \\
Rician &
\cellcolor{datacellgreen}\makecell{$0.909$ \\ {\small $\pm 0.046$}} &
\makecell{$0.882$ \\ {\small $\pm 0.076$}} &
\makecell{$0.894$ \\ {\small $\pm 0.063$}} &
\cellcolor{datacellgreen}\makecell{$0.836$ \\ {\small $\pm 0.074$}} &
\makecell{$0.796$ \\ {\small $\pm 0.114$}} &
\makecell{$0.813$ \\ {\small $\pm 0.084$}} \\
Salt \& Pepper &
\makecell{$0.668$ \\ {\small $\pm 0.195$}} &
\cellcolor{datacellgreen}\makecell{$0.842$ \\ {\small $\pm 0.078$}} &
\makecell{$0.796$ \\ {\small $\pm 0.019$}} &
\makecell{$0.567$ \\ {\small $\pm 0.204$}} &
\cellcolor{datacellgreen}\makecell{$0.743$ \\ {\small $\pm 0.104$}} &
\makecell{$0.689$ \\ {\small $\pm 0.018$}} \\
Speckle &
\makecell{$0.557$ \\ {\small $\pm 0.329$}} &
\cellcolor{datacellgreen}\makecell{$0.836$ \\ {\small $\pm 0.085$}} &
\makecell{$0.787$ \\ {\small $\pm 0.013$}} &
\makecell{$0.465$ \\ {\small $\pm 0.321$}} &
\cellcolor{datacellgreen}\makecell{$0.737$ \\ {\small $\pm 0.113$}} &
\makecell{$0.678$ \\ {\small $\pm 0.011$}} \\
Step Motion MRI &
\makecell{$0.836$ \\ {\small $\pm 0.117$}} &
\makecell{$0.849$ \\ {\small $\pm 0.089$}} &
\cellcolor{datacellgreen}\makecell{$0.888$ \\ {\small $\pm 0.067$}} &
\makecell{$0.733$ \\ {\small $\pm 0.154$}} &
\makecell{$0.748$ \\ {\small $\pm 0.125$}} &
\cellcolor{datacellgreen}\makecell{$0.803$ \\ {\small $\pm 0.085$}} \\
\bottomrule
\end{tabular}
\end{table*}
\subsection{Segmentation Performance Under Box Prompt}
Table~\ref{tab:bbox_main} presents segmentation performance under bounding box prompts across multiple image degradation conditions. Results are reported as the mean $\pm$ standard deviation over different degradation types for each dataset. We compare SAM, MedSAM, and RobustMedSAM using Dice and IoU metrics on six medical imaging datasets. MedSAM consistently outperforms SAM, indicating the effectiveness of medical-domain adaptation. RobustMedSAM further improves robustness under degraded conditions, achieving the best overall Dice score and demonstrating improved stability across challenging datasets.

Figure~\ref{fig:qualitativez} shows qualitative segmentation results under different image degradations using bounding box prompts. Each row corresponds to a representative degradation type, including Gaussian noise, motion artifacts, and contrast variation. From left to right: degraded input with bounding box prompt, ground truth, and predictions from SAM, MedSAM, and RobustMedSAM. Green indicates true positives, red indicates false positives, and blue indicates false negatives. Overall, MedSAM and RobustMedSAM produce more accurate and stable segmentations than SAM under degraded conditions.

%% file: subsection/table_clear_vs_degraded.tex
\begin{table*}[t]
\centering
\caption{Clean vs.\ degraded with \textbf{point prompts} (Dice $\uparrow$). $\Delta$ = degraded $-$ clean; positive values indicate the model benefits from the clear-degraded pair protocol.}
\label{tab:clean_vs_deg}
\resizebox{\linewidth}{!}{
\begin{tabular}{lcccccccccccc}
\toprule
 & \multicolumn{3}{c}{SAM} & \multicolumn{3}{c}{MedSAM} & \multicolumn{3}{c}{RobustMedSAM} & \multicolumn{3}{c}{+SVD} \\
\cmidrule(lr){2-4} \cmidrule(lr){5-7} \cmidrule(lr){8-10} \cmidrule(lr){11-13} 
Dataset & Clean & Deg. & $\Delta$ & Clean & Deg. & $\Delta$ & Clean & Deg. & $\Delta$ & Clean & Deg. & $\Delta$ \\
\midrule
ISIC 2016 & 0.734 & 0.752 & +0.018 & 0.203 & 0.202 & -0.001 & 0.809 & 0.911 & {\color{green!60!black}+0.102} & 0.699 & 0.910 & {\color{green!60!black}+0.210} \\
BUSI & 0.717 & 0.718 & +0.001 & 0.252 & 0.249 & -0.002 & 0.610 & 0.760 & {\color{green!60!black}+0.151} & 0.434 & 0.678 & {\color{green!60!black}+0.243} \\
PROMISE12 & 0.639 & 0.593 & -0.046 & 0.208 & 0.209 & +0.001 & 0.714 & 0.805 & {\color{green!60!black}+0.091} & 0.544 & 0.751 & {\color{green!60!black}+0.207} \\
UWaterloo & 0.840 & 0.841 & +0.001 & 0.201 & 0.202 & +0.002 & 0.736 & 0.869 & {\color{green!60!black}+0.132} & 0.759 & 0.872 & {\color{green!60!black}+0.113} \\
BriFiSeg & 0.321 & 0.311 & -0.010 & 0.112 & 0.114 & +0.002 & 0.289 & 0.305 & +0.016 & 0.125 & 0.436 & {\color{green!60!black}+0.311} \\
AbdomenUS & 0.477 & 0.463 & -0.013 & 0.227 & 0.227 & +0.001 & 0.563 & 0.666 & {\color{green!60!black}+0.103} & 0.261 & 0.612 & {\color{green!60!black}+0.351} \\
\midrule
\textbf{Overall} & 0.621 & 0.613 & \textbf{-0.008} & 0.200 & 0.201 & \textbf{+0.000} & 0.620 & 0.719 & {\color{green!60!black}\textbf{+0.099}} & 0.470 & 0.710 & {\color{green!60!black}\textbf{+0.239}} \\
\bottomrule
\end{tabular}}
\end{table*}

%% file: subsection/table_dice_degradation.tex
\begin{table*}
\centering
\caption{Per-degradation with \textbf{point prompts} — Dice $\uparrow$. Values are mean $\pm$ std across datasets. (Single-dataset degradations show mean only.)}
\label{tab:degradation_dice}
\small
\resizebox{0.7\linewidth}{!}{
\begin{tabular}{lccccc}
\toprule
\makecell[l]{Degradation} &
\makecell[c]{SAM} &
\makecell[c]{MedSAM} &
\makecell[c]{RobustSAM} &
\makecell[c]{Ours - Robust\\MedSAM} &
\makecell[c]{Ours - Robust\\MedSAM+SVD} \\
\midrule

Brightness &
\makecell{$0.619$ \\ {\small $\pm 0.177$}} &
\makecell{$0.206$ \\ {\small $\pm 0.040$}} &
\makecell{$0.633$ \\ {\small $\pm 0.196$}} &
\cellcolor{datacellgreen}\makecell{$0.729$ \\ {\small $\pm 0.195$}} &
\makecell{$0.710$ \\ {\small $\pm 0.166$}} \\

Color Jitter &
\makecell{$0.801$ \\ {\small $\pm 0.059$}} &
\makecell{$0.201$ \\ {\small $\pm 0.017$}} &
\makecell{$0.870$ \\ {\small $\pm 0.043$}} &
\cellcolor{datacellgreen}\makecell{$0.897$ \\ {\small $\pm 0.012$}} &
\makecell{$0.880$ \\ {\small $\pm 0.024$}} \\

Compression &
\makecell{$0.813$ \\ {\small $\pm 0.042$}} &
\makecell{$0.198$ \\ {\small $\pm 0.002$}} &
\makecell{$0.879$ \\ {\small $\pm 0.042$}} &
\makecell{$0.887$ \\ {\small $\pm 0.024$}} &
\cellcolor{datacellgreen}\makecell{$0.905$ \\ {\small $\pm 0.004$}} \\

Contrast &
\makecell{$0.605$ \\ {\small $\pm 0.176$}} &
\makecell{$0.189$ \\ {\small $\pm 0.054$}} &
\makecell{$0.641$ \\ {\small $\pm 0.200$}} &
\cellcolor{datacellgreen}\makecell{$0.717$ \\ {\small $\pm 0.182$}} &
\makecell{$0.696$ \\ {\small $\pm 0.167$}} \\

Gaussian Blur &
\makecell{$0.609$ \\ {\small $\pm 0.163$}} &
\makecell{$0.203$ \\ {\small $\pm 0.050$}} &
\makecell{$0.632$ \\ {\small $\pm 0.206$}} &
\cellcolor{datacellgreen}\makecell{$0.711$ \\ {\small $\pm 0.213$}} &
\makecell{$0.705$ \\ {\small $\pm 0.169$}} \\

Gaussian Noise &
\makecell{$0.625$ \\ {\small $\pm 0.213$}} &
\makecell{$0.206$ \\ {\small $\pm 0.041$}} &
\makecell{$0.639$ \\ {\small $\pm 0.216$}} &
\cellcolor{datacellgreen}\makecell{$0.717$ \\ {\small $\pm 0.230$}} &
\makecell{$0.703$ \\ {\small $\pm 0.182$}} \\

Poisson &
\makecell{$0.321$} &
\makecell{$0.123$} &
\makecell{$0.334$} &
\makecell{$0.345$} &
\cellcolor{datacellgreen}\makecell{$0.472$} \\

Rayleigh &
\makecell{$0.665$} &
\makecell{$0.214$} &
\makecell{$0.563$} &
\cellcolor{datacellgreen}\makecell{$0.816$} &
\makecell{$0.772$} \\

Rician &
\makecell{$0.591$} &
\makecell{$0.203$} &
\makecell{$0.593$} &
\cellcolor{datacellgreen}\makecell{$0.810$} &
\makecell{$0.773$} \\

Salt \& Pepper &
\makecell{$0.582$ \\ {\small $\pm 0.095$}} &
\makecell{$0.257$ \\ {\small $\pm 0.004$}} &
\makecell{$0.525$ \\ {\small $\pm 0.042$}} &
\cellcolor{datacellgreen}\makecell{$0.701$ \\ {\small $\pm 0.056$}} &
\makecell{$0.656$ \\ {\small $\pm 0.038$}} \\

Speckle &
\makecell{$0.594$ \\ {\small $\pm 0.128$}} &
\makecell{$0.217$ \\ {\small $\pm 0.005$}} &
\makecell{$0.602$ \\ {\small $\pm 0.128$}} &
\cellcolor{datacellgreen}\makecell{$0.717$ \\ {\small $\pm 0.041$}} &
\makecell{$0.652$ \\ {\small $\pm 0.049$}} \\

Step Motion MRI &
\makecell{$0.453$} &
\makecell{$0.205$} &
\makecell{$0.464$} &
\cellcolor{datacellgreen}\makecell{$0.779$} &
\makecell{$0.712$} \\

\bottomrule
\end{tabular}}
\end{table*}

%% file: subsection/table_iou_degradation.tex
\begin{table*}
\centering
\caption{Per-degradation analysis with \textbf{point prompts} — IoU $\uparrow$. Values are mean $\pm$ std across datasets. (Single-dataset degradations show mean only.)}
\label{tab:degradation_iou}
\small
\resizebox{0.7\linewidth}{!}{
\begin{tabular}{lccccc}
\toprule
\makecell[l]{Degradation} &
\makecell[c]{SAM} &
\makecell[c]{MedSAM} &
\makecell[c]{RobustSAM} &
\makecell[c]{Ours - Robust\\MedSAM} &
\makecell[c]{Ours - Robust\\MedSAM+SVD} \\
\midrule

Brightness &
\makecell{$0.506$ \\ {\small $\pm 0.190$}} &
\makecell{$0.123$ \\ {\small $\pm 0.027$}} &
\makecell{$0.523$ \\ {\small $\pm 0.219$}} &
\cellcolor{datacellgreen}\makecell{$0.621$ \\ {\small $\pm 0.212$}} &
\makecell{$0.595$ \\ {\small $\pm 0.192$}} \\

Color Jitter &
\makecell{$0.704$ \\ {\small $\pm 0.079$}} &
\makecell{$0.120$ \\ {\small $\pm 0.013$}} &
\makecell{$0.787$ \\ {\small $\pm 0.058$}} &
\cellcolor{datacellgreen}\makecell{$0.820$ \\ {\small $\pm 0.020$}} &
\makecell{$0.799$ \\ {\small $\pm 0.035$}} \\

Compression &
\makecell{$0.709$ \\ {\small $\pm 0.053$}} &
\makecell{$0.118$ \\ {\small $\pm 0.003$}} &
\makecell{$0.802$ \\ {\small $\pm 0.057$}} &
\makecell{$0.809$ \\ {\small $\pm 0.033$}} &
\cellcolor{datacellgreen}\makecell{$0.832$ \\ {\small $\pm 0.010$}} \\

Contrast &
\makecell{$0.494$ \\ {\small $\pm 0.187$}} &
\makecell{$0.113$ \\ {\small $\pm 0.036$}} &
\makecell{$0.532$ \\ {\small $\pm 0.224$}} &
\cellcolor{datacellgreen}\makecell{$0.607$ \\ {\small $\pm 0.199$}} &
\makecell{$0.582$ \\ {\small $\pm 0.193$}} \\

Gaussian Blur &
\makecell{$0.494$ \\ {\small $\pm 0.175$}} &
\makecell{$0.120$ \\ {\small $\pm 0.033$}} &
\makecell{$0.522$ \\ {\small $\pm 0.227$}} &
\cellcolor{datacellgreen}\makecell{$0.605$ \\ {\small $\pm 0.221$}} &
\makecell{$0.591$ \\ {\small $\pm 0.193$}} \\

Gaussian Noise &
\makecell{$0.519$ \\ {\small $\pm 0.226$}} &
\makecell{$0.123$ \\ {\small $\pm 0.027$}} &
\makecell{$0.532$ \\ {\small $\pm 0.235$}} &
\cellcolor{datacellgreen}\makecell{$0.615$ \\ {\small $\pm 0.232$}} &
\makecell{$0.589$ \\ {\small $\pm 0.201$}} \\

Poisson &
\makecell{$0.195$} &
\makecell{$0.068$} &
\makecell{$0.205$} &
\makecell{$0.224$} &
\cellcolor{datacellgreen}\makecell{$0.324$} \\

Rayleigh &
\makecell{$0.552$} &
\makecell{$0.127$} &
\makecell{$0.441$} &
\cellcolor{datacellgreen}\makecell{$0.703$} &
\makecell{$0.647$} \\

Rician &
\makecell{$0.480$} &
\makecell{$0.120$} &
\makecell{$0.460$} &
\cellcolor{datacellgreen}\makecell{$0.693$} &
\makecell{$0.645$} \\

Salt \& Pepper &
\makecell{$0.458$ \\ {\small $\pm 0.109$}} &
\makecell{$0.160$ \\ {\small $\pm 0.006$}} &
\makecell{$0.401$ \\ {\small $\pm 0.055$}} &
\cellcolor{datacellgreen}\makecell{$0.571$ \\ {\small $\pm 0.069$}} &
\makecell{$0.526$ \\ {\small $\pm 0.049$}} \\

Speckle &
\makecell{$0.473$ \\ {\small $\pm 0.140$}} &
\makecell{$0.131$ \\ {\small $\pm 0.004$}} &
\makecell{$0.479$ \\ {\small $\pm 0.140$}} &
\cellcolor{datacellgreen}\makecell{$0.584$ \\ {\small $\pm 0.054$}} &
\makecell{$0.521$ \\ {\small $\pm 0.059$}} \\

Step Motion MRI &
\makecell{$0.341$} &
\makecell{$0.121$} &
\makecell{$0.338$} &
\cellcolor{datacellgreen}\makecell{$0.652$} &
\makecell{$0.577$} \\

\bottomrule
\end{tabular}}
\end{table*}

%% file: subsection/table_nsd_degradation.tex
\begin{table*}
\centering
\caption{Per-degradation analysis with \textbf{point prompts} — NSD $\uparrow$ ($\tau{=}2$\,px). Values are mean $\pm$ std across datasets. (Single-dataset degradations show mean only.)}
\label{tab:degradation_nsd}
\small
\resizebox{0.7\linewidth}{!}{
\begin{tabular}{lccccc}
\toprule
\makecell[l]{Degradation} &
\makecell[c]{SAM} &
\makecell[c]{MedSAM} &
\makecell[c]{RobustSAM} &
\makecell[c]{Ours - Robust\\MedSAM} &
\makecell[c]{Ours - Robust\\MedSAM+SVD} \\
\midrule

Brightness &
\makecell{$0.191$ \\ {\small $\pm 0.088$}} &
\makecell{$0.036$ \\ {\small $\pm 0.011$}} &
\cellcolor{datacellgreen}\makecell{$0.203$ \\ {\small $\pm 0.112$}} &
\makecell{$0.199$ \\ {\small $\pm 0.066$}} &
\makecell{$0.182$ \\ {\small $\pm 0.053$}} \\

Color Jitter &
\makecell{$0.212$ \\ {\small $\pm 0.138$}} &
\makecell{$0.038$ \\ {\small $\pm 0.021$}} &
\cellcolor{datacellgreen}\makecell{$0.274$ \\ {\small $\pm 0.149$}} &
\makecell{$0.262$ \\ {\small $\pm 0.028$}} &
\makecell{$0.245$ \\ {\small $\pm 0.022$}} \\

Compression &
\makecell{$0.228$ \\ {\small $\pm 0.107$}} &
\makecell{$0.030$ \\ {\small $\pm 0.012$}} &
\cellcolor{datacellgreen}\makecell{$0.296$ \\ {\small $\pm 0.133$}} &
\makecell{$0.275$ \\ {\small $\pm 0.023$}} &
\makecell{$0.252$ \\ {\small $\pm 0.019$}} \\

Contrast &
\makecell{$0.184$ \\ {\small $\pm 0.092$}} &
\makecell{$0.036$ \\ {\small $\pm 0.011$}} &
\cellcolor{datacellgreen}\makecell{$0.209$ \\ {\small $\pm 0.127$}} &
\makecell{$0.205$ \\ {\small $\pm 0.069$}} &
\makecell{$0.195$ \\ {\small $\pm 0.062$}} \\

Gaussian Blur &
\makecell{$0.178$ \\ {\small $\pm 0.080$}} &
\makecell{$0.037$ \\ {\small $\pm 0.010$}} &
\makecell{$0.195$ \\ {\small $\pm 0.105$}} &
\cellcolor{datacellgreen}\makecell{$0.202$ \\ {\small $\pm 0.066$}} &
\makecell{$0.194$ \\ {\small $\pm 0.060$}} \\

Gaussian Noise &
\makecell{$0.190$ \\ {\small $\pm 0.116$}} &
\makecell{$0.038$ \\ {\small $\pm 0.012$}} &
\cellcolor{datacellgreen}\makecell{$0.207$ \\ {\small $\pm 0.141$}} &
\makecell{$0.199$ \\ {\small $\pm 0.069$}} &
\makecell{$0.193$ \\ {\small $\pm 0.058$}} \\

Poisson &
\makecell{$0.108$} &
\makecell{$0.051$} &
\makecell{$0.104$} &
\makecell{$0.143$} &
\cellcolor{datacellgreen}\makecell{$0.193$} \\

Rayleigh &
\makecell{$0.192$} &
\makecell{$0.029$} &
\makecell{$0.132$} &
\cellcolor{datacellgreen}\makecell{$0.218$} &
\makecell{$0.188$} \\

Rician &
\makecell{$0.170$} &
\makecell{$0.028$} &
\makecell{$0.132$} &
\cellcolor{datacellgreen}\makecell{$0.206$} &
\makecell{$0.189$} \\

Salt \& Pepper &
\makecell{$0.153$ \\ {\small $\pm 0.041$}} &
\makecell{$0.041$ \\ {\small $\pm 0.003$}} &
\makecell{$0.165$ \\ {\small $\pm 0.002$}} &
\cellcolor{datacellgreen}\makecell{$0.170$ \\ {\small $\pm 0.061$}} &
\makecell{$0.157$ \\ {\small $\pm 0.058$}} \\

Speckle &
\makecell{$0.176$ \\ {\small $\pm 0.053$}} &
\makecell{$0.041$ \\ {\small $\pm 0.002$}} &
\cellcolor{datacellgreen}\makecell{$0.215$ \\ {\small $\pm 0.022$}} &
\makecell{$0.165$ \\ {\small $\pm 0.061$}} &
\makecell{$0.154$ \\ {\small $\pm 0.066$}} \\

Step Motion MRI &
\makecell{$0.091$} &
\makecell{$0.027$} &
\makecell{$0.078$} &
\cellcolor{datacellgreen}\makecell{$0.164$} &
\makecell{$0.155$} \\

\bottomrule
\end{tabular}}
\end{table*}

%% file: main.bib
@inproceedings{kirillov2023segment,
  title={Segment anything},
  author={Kirillov, Alexander and Mintun, Eric and Ravi, Nikhila and Mao, Hanzi and Rolland, Chloe and Gustafson, Laura and Xiao, Tete and Whitehead, Spencer and Berg, Alexander C and Lo, Wan-Yen and others},
  booktitle={Proceedings of the IEEE/CVF international conference on computer vision},
  pages={4015--4026},
  year={2023}
}

@article{zhang2023enhancing,
  title={Enhancing the reliability of segment anything model for auto-prompting medical image segmentation with uncertainty rectification},
  author={Zhang, Yichi and Hu, Shiyao and Ren, Sijie and Jiang, Chen and Cheng, Yuan and Qi, Yuan},
  journal={arXiv preprint arXiv:2311.10529},
  year={2023}
}

@article{noise,
  title={Noise issues prevailing in various types of medical images},
  author={Goyal, Bhawna and Agrawal, Sunil and Sohi, BS},
  journal={Biomedical \& Pharmacology Journal},
  volume={11},
  number={3},
  pages={1227},
  year={2018},
  publisher={Biomedical and Pharmacology Journal}
}

@article{ma2024segment,
  title={Segment anything in medical images},
  author={Ma, Jun and He, Yuting and Li, Feifei and Han, Lin and You, Chenyu and Wang, Bo},
  journal={Nature Communications},
  volume={15},
  number={1},
  pages={654},
  year={2024},
  publisher={Nature Publishing Group UK London}
}

@article{shaharabany2023autosam,
  title={Autosam: Adapting sam to medical images by overloading the prompt encoder},
  author={Shaharabany, Tal and Dahan, Aviad and Giryes, Raja and Wolf, Lior},
  journal={arXiv preprint arXiv:2306.06370},
  year={2023}
}

@InProceedings{Pandey_2023_ICCV,
    author    = {Pandey, Sumit and Chen, Kuan-Fu and Dam, Erik B.},
    title     = {Comprehensive Multimodal Segmentation in Medical Imaging: Combining YOLOv8 with SAM and HQ-SAM Models},
    booktitle = {Proceedings of the IEEE/CVF International Conference on Computer Vision (ICCV) Workshops},
    month     = {October},
    year      = {2023},
    pages     = {2592-2598}
}

@article{medlsam,
  title        = {{MedLSAM}: Localize and Segment Anything Model for 3D CT Images},
  author       = {Lei, Wenhui and Xu, Wei and Li, Kang and Zhang, Xiaofan and Zhang, Shaoting},
  journal      = {Medical Image Analysis},
  volume       = {99},
  pages        = {103370},
  year         = {2025},
  issn         = {1361-8415},
  doi          = {10.1016/j.media.2024.103370},
  url          = {https://doi.org/10.1016/j.media.2024.103370},
}

@inproceedings{yue2024surgicalsam,
  title={Surgicalsam: Efficient class promptable surgical instrument segmentation},
  author={Yue, Wenxi and Zhang, Jing and Hu, Kun and Xia, Yong and Luo, Jiebo and Wang, Zhiyong},
  booktitle={Proceedings of the AAAI Conference on Artificial Intelligence},
  volume={38},
  number={7},
  pages={6890--6898},
  year={2024}
}

@inproceedings{tejero2025sam,
  title={Sam-da: Decoder adapter for efficient medical domain adaptation},
  author={Tejero, Javier Gamazo and Schmid, Moritz J and Neila, Pablo M{\'a}rquez and Zinkernagel, Martin and Wolf, Sebastian and Sznitman, Raphael},
  booktitle={Proceedings of the Winter Conference on Applications of Computer Vision},
  pages={6775--6784},
  year={2025}
}

@inproceedings{chen2024robustsam,
  title={Robustsam: Segment anything robustly on degraded images},
  author={Chen, Wei-Ting and Vong, Yu-Jiet and Kuo, Sy-Yen and Ma, Sizhou and Wang, Jian},
  booktitle={Proceedings of the IEEE/CVF Conference on Computer Vision and Pattern Recognition},
  pages={4081--4091},
  year={2024}
}

@inproceedings{yang2020fda,
  title={Fda: Fourier domain adaptation for semantic segmentation},
  author={Yang, Yanchao and Soatto, Stefano},
  booktitle={Proceedings of the IEEE/CVF conference on computer vision and pattern recognition},
  pages={4085--4095},
  year={2020}
}

@article{kucs2024medsegbench,
  title={MedSegBench: A comprehensive benchmark for medical image segmentation in diverse data modalities},
  author={Ku{\c{s}}, Zeki and Aydin, Musa},
  journal={Scientific Data},
  volume={11},
  number={1},
  pages={1283},
  year={2024},
  publisher={Nature Publishing Group UK London}
}

@inproceedings{xiong2024efficientsam,
  title={Efficientsam: Leveraged masked image pretraining for efficient segment anything},
  author={Xiong, Yunyang and Varadarajan, Bala and Wu, Lemeng and Xiang, Xiaoyu and Xiao, Fanyi and Zhu, Chenchen and Dai, Xiaoliang and Wang, Dilin and Sun, Fei and Iandola, Forrest and others},
  booktitle={Proceedings of the IEEE/CVF conference on computer vision and pattern recognition},
  pages={16111--16121},
  year={2024}
}

@inproceedings{cheng2024unleashing,
  title={Unleashing the potential of sam for medical adaptation via hierarchical decoding},
  author={Cheng, Zhiheng and Wei, Qingyue and Zhu, Hongru and Wang, Yan and Qu, Liangqiong and Shao, Wei and Zhou, Yuyin},
  booktitle={Proceedings of the IEEE/CVF conference on computer vision and pattern recognition},
  pages={3511--3522},
  year={2024}
}

@inproceedings{kamann2020benchmarking,
  title={Benchmarking the robustness of semantic segmentation models},
  author={Kamann, Christoph and Rother, Carsten},
  booktitle={Proceedings of the IEEE/CVF conference on computer vision and pattern recognition},
  pages={8828--8838},
  year={2020}
}

@inproceedings{rajagopalan2023improving,
  title={Improving robustness of semantic segmentation to motion-blur using class-centric augmentation},
  author={Rajagopalan, AN and others},
  booktitle={Proceedings of the IEEE/CVF Conference on Computer Vision and Pattern Recognition},
  pages={10470--10479},
  year={2023}
}

@inproceedings{peng2024parameter,
  title={Parameter efficient fine-tuning via cross block orchestration for segment anything model},
  author={Peng, Zelin and Xu, Zhengqin and Zeng, Zhilin and Xie, Lingxi and Tian, Qi and Shen, Wei},
  booktitle={Proceedings of the IEEE/CVF Conference on Computer Vision and Pattern Recognition},
  pages={3743--3752},
  year={2024}
}

@inproceedings{wu2024one,
  title={One-prompt to segment all medical images},
  author={Wu, Junde and Xu, Min},
  booktitle={Proceedings of the IEEE/CVF conference on computer vision and pattern recognition},
  pages={11302--11312},
  year={2024}
}

@inproceedings{zhang2024improving,
  title={Improving the generalization of segmentation foundation model under distribution shift via weakly supervised adaptation},
  author={Zhang, Haojie and Su, Yongyi and Xu, Xun and Jia, Kui},
  booktitle={Proceedings of the IEEE/CVF Conference on Computer Vision and Pattern Recognition},
  pages={23385--23395},
  year={2024}
}

@inproceedings{nam2024modality,
  title={Modality-agnostic domain generalizable medical image segmentation by multi-frequency in multi-scale attention},
  author={Nam, Ju-Hyeon and Syazwany, Nur Suriza and Kim, Su Jung and Lee, Sang-Chul},
  booktitle={Proceedings of the IEEE/CVF conference on computer vision and pattern recognition},
  pages={11480--11491},
  year={2024}
}

@inproceedings{chen2024versatile,
  title={Versatile medical image segmentation learned from multi-source datasets via model self-disambiguation},
  author={Chen, Xiaoyang and Zheng, Hao and Li, Yuemeng and Ma, Yuncong and Ma, Liang and Li, Hongming and Fan, Yong},
  booktitle={Proceedings of the IEEE/CVF Conference on Computer Vision and Pattern Recognition},
  pages={11747--11756},
  year={2024}
}
